\documentclass[preprint,12pt]{elsarticle}
\usepackage{picins}
\usepackage{amssymb}
\usepackage{float}
\usepackage{amsmath}
\usepackage{algorithm}
\usepackage[misc]{ifsym} 
\usepackage{rotating}

\usepackage{color, soul, framed}
\usepackage{color,xcolor}
\usepackage{subfig}
\usepackage{subfloat}
\usepackage{url}
\usepackage{hyperref}
\RequirePackage{pdflscape}
\usepackage{algpseudocode}
\journal{Neurocomputing}
\usepackage{threeparttable}
\begin{document}
	
	\begin{frontmatter}

		\title{Causal and Local Correlations Based Network for Multivariate Time Series Classification}
		
		\author[a]{Mingsen Du} \ead{MingsenDu@163.com}
		\author[a]{Yanxuan Wei} \ead{wyx\_num1@163.com}
		\author[a,b]{Xiangwei Zheng} \ead{xwzhengcn@163.com}
		\author[a,b]{Cun Ji\corref{cor1}}  \ead{jicun@sdnu.edu.cn}
		
		\cortext[cor1]{Corresponding author}
		\address[a]{School of Information Science and Engineering, Shandong Normal University, Jinan, China}
		\address[b]{Shandong Provincial Key Laboratory for Distributed Computer Software Novel Technology, Jinan, China}
		
		\begin{abstract}
			Recently, time series classification has attracted the attention of a large number of researchers, and hundreds of methods have been proposed. However, these methods often ignore the spatial correlations among dimensions and the local correlations among features. To address this issue, the causal and local correlations based network (CaLoNet) is proposed in this study for multivariate time series classification. First, pairwise spatial correlations between dimensions are modeled using causality modeling to obtain the graph structure. Then, a relationship extraction network is used to fuse local correlations to obtain long-term dependency features. Finally, the graph structure and long-term dependency features are integrated into the graph neural network. Experiments on the UEA datasets show that CaLoNet can obtain competitive performance compared with state-of-the-art methods.
		\end{abstract}

	\end{frontmatter}
	
	\section{Introduction}
	
	With the development of the Internet of Things, large numbers of sensors are now used to periodically collect data. These collected data naturally exist in the form of time series. Usually, time series contain multiple variables with correlations. In other words, the collected data are in the format of a multivariate time series (MTS). MTS classification aims to assign predefined labels to time series using supervised learning \cite{ji2022time}, and it is widely used in various real-life domains, such as human behavior recognition \cite{ma2019attnsense}, healthcare \cite{strodthoff2019detecting, Li2024AnAF}, macroeconomics \cite{cochrane2005time}, and misinformation detection \cite{yu2022lstm}.
	
	In recent years, many different methods based on different patterns have been proposed. Among them, methods based on various correlations have been widely explored \cite{Zuo2023SVPTAS, Liu2023TodyNetTD}. The variables in MTS data usually exhibit complex correlations, and fully using these correlations can help build more accurate prediction models. Therefore, ignoring the correlations between variables leads to information loss and reduces prediction performance. By analyzing the local and dimensional correlations in MTS, we can discover hidden patterns, regularities, and potential structures within the data, thereby gaining a deeper understanding and insight. The correlations between variables may reflect certain causal mechanisms, providing clues for uncovering the underlying mechanisms that generate the data. This holds particular significance in fields such as healthcare, finance, and meteorology, as it helps unravel the intrinsic connections among phenomena.
	
\begin{figure}[!tb]
		\centering
		\includegraphics[width=0.8\textwidth]{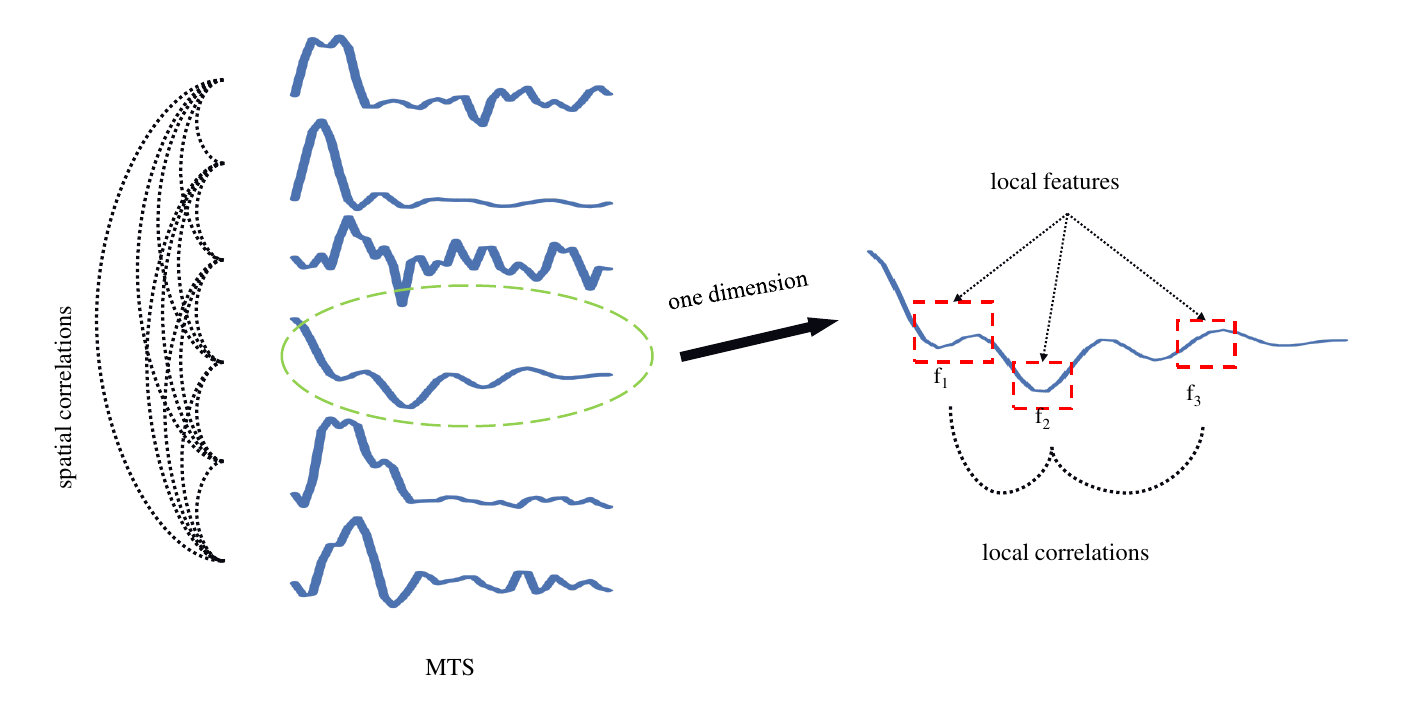}
		\caption{Examples of spatial correlations and local correlations in MTSs.}
		\label{fig:fig1}
	\end{figure}

	MTSs typically exhibit the following correlations: 1) \textbf{Local correlations} within a single dimension. Subsequent observations in a time series are influenced by the preceding observations. For example, future traffic flow is influenced by current traffic flow, and specific streets are more likely to be influenced by traffic information from neighboring areas. The right side of Fig.~\ref{fig:fig1} presents an example of local correlations of local features (i.e., shapelets \cite{ye2009time}, indicated by the dashed red boxes). 2) \textbf{Spatial correlations} among dimensions. Each variable in an MTS is influenced by other variables. Some studies \cite{duan2022multivariate,cao2020spectral} assume that the predicted values of individual variables are influenced by all other variables or exhibit other spatial correlations. The left side of Fig.~\ref{fig:fig1} depicts spatial correlations. Considering this information is highly beneficial for exploring the interactions among MTSs.
	
	\textbf{Regarding local correlations}, Transformer-based methods have shown great potential in recent years, and numerous methods have used this model for MTS tasks \cite{Chen2024EliminationOR}. A large number of methods have been proposed to extract local correlations \cite{Bahri2022ShapeletbasedTA}. The authors of \cite{Zuo2023SVPTAS} proposed a variable-position Transformer to extract local correlations between variables. It takes time series subsequences, which can come from different variables and positions (time intervals), as input (at shape level). The local correlations capture long- and short-term dependencies between shapes to process them. In addition, other methods have improved the Transformer and made progress in extracting local correlation \cite{Yang2023DyformerAD, Xiao2024DenselyKN, Wang2022EnhancingTE, Cheng2023FormerTimeHM, Foumani2023ImprovingPE}. These methods have contributed to improving the capabilities of modeling time series data and processing local correlations.
	
	\textbf{Regarding spatial correlations}, numerous methods have been proposed, among which methods based on the self-attention mechanism (Transformer) and graph neural networks (GNNs) have received much attention. Methods based on the self-attention mechanism include a variable-position Transformer proposed by \cite{Zuo2023SVPTAS} to extract correlations between dimensions. The authors of \cite{Chen2022CaSSAC} designed a channel-aware Transformer encoder to capture the complex relationships between different time channels in an MTS. However, the Transformer has the following major drawbacks when extracting spatial correlations from an MTS compared with a GNN: (1) Lack of explicit structural modeling capability. The Transformer relies on a fully connected self-attention mechanism to learn the correlations between sequences, but cannot explicitly model the topological relationships between variables. A GNN, by contrast, can directly operate on the defined graph structure, more easily capturing and using the explicit relationships among variables. (2) High computational complexity. The Transformer's self-attention mechanism must compute the correlation scores between all element pairs, resulting in a computational complexity of $O(n^2)$. On a high-dimensional MTS, the Transformer may encounter computational bottlenecks. A GNN, however, can effectively use sparse connections to reduce computational overhead. (3) Difficulty in using prior knowledge. The Transformer mainly learns the relationships between variables in a data-driven manner, and lacks mechanisms for using prior domain knowledge. In many application scenarios, however, we may already know some of the topological structures between variables, which could help simplify the learning task of a GNN. (4) Lack of reasoning ability on graphs. The Transformer learns implicit variable correlation representations and lacks the ability to reason on explicit graph structures like a GNN, which complicates interpretability analysis.

	Methods based on GNN and graph learning include the time dynamic GNN \cite{Liu2023TodyNetTD}, which can extract hidden spatial-temporal dependencies without defining the graph structure and includes a time graph pooling layer to obtain a global graph-level representation of graph learning with learnable time parameters. Graph learning \cite{Wang2023IrregularlySM}, \cite{Wang2023FullyConnectedSG}, \cite{Wang2024MultivariateTR} provides greater flexibility, but also leads to higher computational complexity, uncertainty, and data demands. Compared with using a fixed predefined graph structure, graph learning methods have the following drawbacks: (1) Higher computational complexity. Graph learning methods need to simultaneously optimize the graph structure and model parameters, which usually significantly increases computational costs. Fixed graph structures, by contrast, avoid the overhead of learning the graph structure. (2) Poor convergence and stability. Because the graph structure and parameters are optimized simultaneously, graph learning problems are often non-convex, prone to suboptimal solutions, and have poor convergence and stability. In contrast, training models on a fixed graph structure is more stable. (3) Lack of theoretical guarantees. Current graph learning methods are mainly based on empirical or heuristic algorithms, which lack theoretical convergence guarantees and performance bounds, making it difficult to ensure that the learned graph structure is optimal. In some traditional tasks or scenarios with limited data, predefined fixed graph structures may be more suitable.
	
	Overall, the Transformer is more suitable for modeling global correlations, whereas the GNN excels at capturing and using structured prior knowledge, and the two models have some complementarity. Combining the advantages of these two types of models has the potential to better exploit the rich information in MTS data.

	However, current research still faces the following challenges: (1)\textbf{ Modeling the spatial correlations between MTS dimensions does not lead to explicit representations}. Moreover, modeling in an unreasonable way may lead to indistinguishable representations, resulting in poor accuracy. The Transformer \cite{Zuo2023SVPTAS} relies on a fully connected self-attention mechanism to learn the correlations between sequences, but cannot explicitly model the topological relationships between variables. The GNN \cite{Liu2023TodyNetTD}, by contrast, can directly operate on the defined graph structure or the learned graph structure, making it easier to capture and use the explicit relationships between variables. Compared with using a fixed predefined graph structure, graph learning methods \cite{Liu2023TodyNetTD, Wang2023IrregularlySM} have drawbacks such as higher computational complexity and poor convergence and stability.
	(2) \textbf{Various MTS classification methods based on local correlations are often unable to more effectively and efficiently model local correlations and incorporate them into representation learning}.
	Compared with other local correlation modeling methods, the Transformer has some unique advantages in extracting local correlations. For example, a Transformer based on the self-attention mechanism can directly capture the dependencies between any two time steps in the sequence, regardless of the time step distance. This allows it to effectively learn long-range local correlation patterns. However, the Transformer's self-attention mechanism must compute the correlation scores between all element pairs in the sequence, resulting in a high (quadratic) computational complexity. This is a challenge for long sequences and high-dimensional sequences. Extensive research \cite{Yang2023DyformerAD, Wang2022EnhancingTE, Cheng2023FormerTimeHM} has been conducted on reducing complexity issues.
	
	To address these challenges, this work proposes a novel end-to-end deep-learning model called the causal and local correlations based network (CaLoNet). In CaLoNet, the first step is to leverage causal correlations to model the pairwise spatial correlations between the dimensions of an MTS using graph structures. Next, a relationship extraction network is employed to fuse local correlations, enabling the extraction of long-term dependency features. Finally, the graph structures and long-term dependency features are incorporated into a GNN.
	
	The main contributions of this paper are summarized as follows.
	\begin{enumerate}
		\item A novel MTS classification network is designed to exploit spatial and local correlations. In this network, spatial correlations and local correlations are jointly used to model the time series for MTS classification.
		\item A novel strategy for constructing the graph structure of an MTS is proposed. This strategy characterizes the causal information through transfer entropy. Additionally, we further propose a graph construction strategy that serves as a new form of time series representation for graph-level classification tasks in MTS classification.
		\item We designed a novel representation learning approach for time series spatial correlations in MTS classification. Graph-level tasks include graph classification, graph regression, and graph matching, all of which require models to learn graph representations.
		\item A large number of experiments on publicly available datasets demonstrate the effectiveness of our method.
	\end{enumerate}
	
	The remainder of this paper is structured as follows. Section~\ref{sec:rel} introduces the related work. Section~\ref{sec3} describes our method in detail.
	The experimental results are presented in Section~\ref{sec:exp}, and our conclusions are provided in Section~\ref{sec:con}.
	
	\section{Related work}
	\label{sec:rel}
	
	\subsection{MTS classification methods}
	MTS classification aims to predict the class label of an unlabeled time series as accurately as possible.
	Recently, many research studies have been devoted to MTS classification, and they have proposed a large number of MTS classification methods. These methods can be divided into traditional machine learning-based methods and deep-learning--based methods.  
	
	Traditional machine learning-based methods usually extract features through tedious feature engineering or data preprocessing. Many machine learning-based methods have been proposed, and they can be generally classified into two categories \cite{ruiz2021great}: 1) distance-based methods, which use the distance between time series as a similarity metric \cite{abanda2019review}. For example, G\'{o}recki et al. \cite{gorecki2015multivariate} proposed PDDTW , which combines the dynamic time warping (DTW) distance between MTS with the DTW distance between derivatives of MTS for classification. 2) Feature-based methods classify time series based on several features \cite{ji2022time,ji2022fully} such as structural features and statistical features. Sch{\"a}fer and Leser \cite{schafer2017multivariate} proposed WEASEL+MUSE, which uses a bag-of-symbolic-Fourier-approximations model for MTS classification. Baydogan and Runger \cite{baydogan2016time} introduced SMTS, which uses a code book to capture the local relationships between different dimensions for MTS classification. 
	
	Deep-learning--based methods aim to unfold the internal representational hierarchy of time series, which helps to capture the intrinsic correlations among the representations \cite{lecun2015deep}. In contrast to traditional machine learning-based methods, deep-learning--based methods successfully solve the problem of mining raw low-dimensional time series to create high-dimension features, and feature extraction can be performed in an end-to-end form.
	
	Consequently, this study focuses on deep-learning--based MTS classification methods because of their excellent ability to capture relationships. Several representative deep-learning--based methods are described in following subsection.
	
	\subsection{Deep-learning--based methods}
	In recent years, increasingly more researchers have classified MTSs through deep learning. Chen et al. \cite{chen2021time} used a multi-level discrete wavelet decomposition to decompose an MTS into a group of sub-MTSs to extract multi-level time-frequency representations. In \cite{chen2021time}, a convolutional neural network (CNN) was developed for each level to learn level-specific nonlinear features, and a metric learning layer was added on the top of the network to learn the semantic similarity of MTSs. Zheng et al. \cite{zheng2014time} first learned features from individual time series in each channel, and then combined features from all channels to achieve classification. Tripathi and Baruah  \cite{tripathi2020multivariate} used an attention-based CNN to encode information across multiple time stamps to learn the temporal features of an MTS. Huang et al. \cite{huang2021functional} proposed FDESN, which is a novel bi-level approach for optimizing parameters. FDESN uses temporal and spatial aggregations for MTS classification. Zhang et al. \cite{zhang2020tapnet} designed a random group permutation method combined with a CNN to learn the features. 
	
	In an MTS, some correlations among features exist. Moreover, the local attributes in the time series classification problem are ordered, which makes this task clearly different from traditional classification problems. In essence, it is irrelevant whether the attributes in time series are ordered in time or not; what matters is the possible presence of order-dependent discriminative features in time series \cite{xiao2021rtfn}. It is also crucial to discover the intrinsic correlations among features extracted from different positions. Various deep-learning methods have their own priorities, and we focus on two methods that are relevant to this study in the two following subsections.
	
	\subsection{Spatial and causal correlation-based methods}
	There is a certain relationship between different dimensions of an MTS. Spatial correlations can be viewed as dependencies between MTSs, and they can be obtained by several quantitative methods. Zuo et al. \cite{zuo2021smate} modeled spatial-temporal dynamic features to demonstrate that the temporal dependency and evolution of the spatial interactions are important for MTS classification. However, this method does not have an explicit spatial structure. 
	
	Causal correlations between MTS dimensions can be viewed as a subset of spatial correlations. Yang et al. \cite{yang2017granger} proposed a method that approximates the time series dynamics and explicitly learns the causal correlation-based relationships among multiple variables. Duan et al. \cite{duan2022multivariate} combined a GNN and an encoder-decoder-based variational graph pooling module to create adaptive centroids for graph coarsening. Zha et al. \cite{zha2022towards} represented time series classification as a node-level classification problem in a graph, where nodes correspond to each time series and links correspond to similarities between time series that are used in the final time series classification.
	
	\subsection{Local correlation-based methods}
	Local correlation-based methods usually contain a feature extraction network and a relationship network \cite{xiao2021rtfn}. The relationship network focuse on extracting the relationships among the features that are obtained by the feature extraction network.  The relationship networks can be a Transformer-based model, a self-attention--based model or a long short-term memory (LSTM)-based model.   
	
	The model proposed by Xiao et al. \cite{xiao2021rtfn} included a temporal feature network to extract local features extraction and an LSTM-based attention network to mine the intrinsic correlations among the features. Liu et al. \cite{liu2021gated} proposed the GTN, an extension of the current Transformer with a gate. The GTN can model channel-wise and step-wise correlations individually. Karim et al. \cite{karim2019multivariate} used an LSTM and a FCN with a squeeze-and-excitation (SE) block \cite{hu2018squeeze} to capture feature correlations. Chen et al. \cite{chen2022net} used a sparse self-attention (SSA)-based network to extract local features and correlations. Hao et al. \cite{hao2020new} introduced a temporal attention mechanism to extract long- and short-term dependencies cross all time steps. Yu et al. \cite{yu2022lstm} designed a traffic incident classifier based on an LSTM. This classifier  was trained on time series feature vectors from both normal and collusion attack scenarios, and hence it can recognized dynamic traffic parameter patterns. Hong et al. \cite{hong2022long} proposed LMGRU, which can obtain the local correlations of time series effectively. Zhang et al. \cite{zhang2023attention} used a temporal attention encoder for extracting global temporal features and convolution for extracting local temporal features. Their method proved the effectiveness of hybrid global--local temporal attention features for MTS classification. 
	
	\section{Proposed method}
	\label{sec3}
	In this section, we first briefly introduce the overall structure of CaLoNet, and then detail the local correlation network, causal correlation network, and the specifics of the completed classification task.
	
	\subsection{Overview}
\begin{figure}[!tb]
		\centering
		\includegraphics[width=0.95\textwidth]{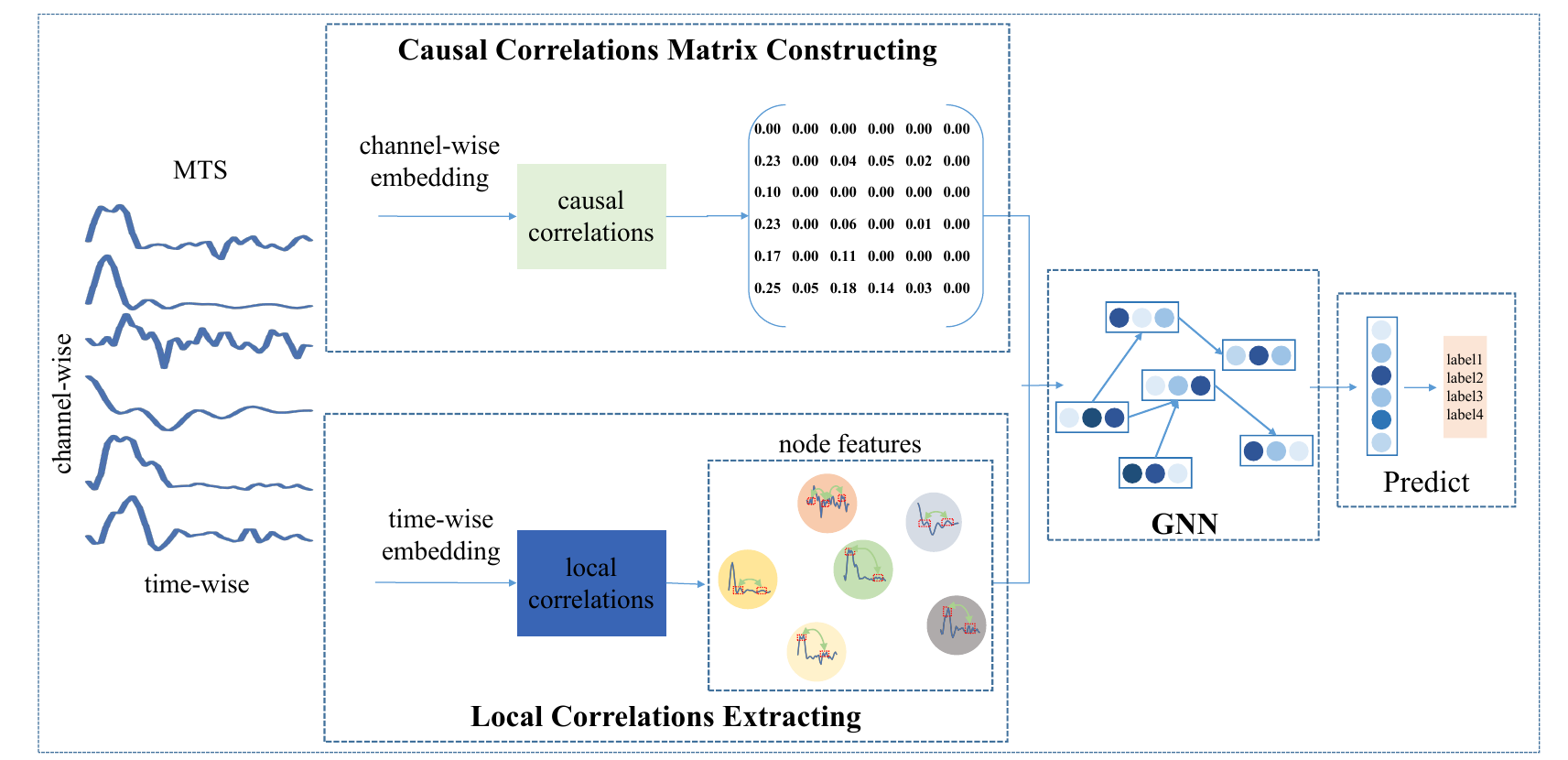}
		\caption{Overall structure of CaLoNet.}
		\label{fig:fig2}
	\end{figure}

	The overall architecture of CaLoNet is shown in Fig.~\ref{fig:fig2}. CaLoNet consists of four main steps:
	\begin{itemize}
		\item \textbf{Causal correlation matrix construction}. As shown in the upper left of Fig.~\ref{fig:fig2}, this step obtains the causal graph matrix between dimensions with the help of transfer entropy, as detailed in Section~\ref{sec3.2}. 
		\item \textbf{Local correlation extraction}. As shown in the bottom left of Fig.~\ref{fig:fig2}, this step obtains the correlations among local features through a local correlation network. Each colored dot represents the local correlation features of each dimension, as detailed in Section~\ref{sec3.3}.
		\item \textbf{Node embedding}. This step obtains the node embedding using a GNN based on the causal graph matrix and node features of the local correlations, as presented in Section~\ref{sec3.4}. 
		\item \textbf{Prediction}. This step predicts the class labels using a multi-layer perceptron (MLP), as described in Section~\ref{sec3.5}.  
	\end{itemize}
	
	\begin{algorithm} 	
		
		\caption{CaLoNet}
		\label{alg1}
		\footnotesize
		\begin{algorithmic}[1]
			\Require training set: $SET$, epochs: $epoch$, each sample in $SET$: T;
			\Ensure  CaLoNet classifier: $CaLoNet$;	
			
			\Function {Cc} {T}
			\For{$ i, j$ in $T$}
			\State $C_{T_{i}, T_{j} } = TE_{T_{i}\rightarrow T_{j}} - TE_{T_{j}\rightarrow T_{i}}$	
			\Comment Transfer entropy
			\If{$C_{T_{i}, T_{j} }>c$}
			\State	$M_{i,j}$=$C_{T_{i},T_{j}}$
			\Else
			\State $M_{i,j}$=0
			\EndIf
			\EndFor
			\State \Return{ $M$ } 
			\EndFunction  
			\\
			\State $M$ = $Cc$($T$)
			\Comment Step 1: Causal correlation matrix construction
			\State $LC$ = $SSA$($T$)
			\Comment Step 2: Local correlation extraction
			
			\State $NE=GIN$($M$, $LC$)
			\Comment Step 3: Obtains node embedding based on the causal graph matrix and node features based on local correlations
			
			\State $CaLoNet$=training($SET$, $NE$, $epoch$)
			\Comment Step 4: CaLoNet training 
			
			\State \Return{$CaLoNet$}
		\end{algorithmic}
	\end{algorithm}
	
	\subsection{Causal correlation matrix construction}
	\label{sec3.2}
	
	Transfer entropy is used to calculate how much information is reduced in an observed system, and CaLoNet constructs the causal-based spatial correlation graph with the help of transfer entropy.
	
	Granger causality analysis \cite{granger1969investigating, seth2007granger} is one of the best-known methods for quantitatively characterizing time series causality. However, as a linear model, Granger causality analysis cannot handle the possible nonlinear correlations of an MTS well. Therefore, transfer entropy \cite{schreiber2000measuring} was proposed to perform a causality analysis that can handle the nonlinear cases. The study \cite{schreiber2000measuring} first introduced transfer entropy as an information-theoretic-based causality measure. The transfer entropy from dimension $Y$ to dimension $X$ is defined in the following equation. In this equation, $x_{t}$ and $y_{t}$ denote the values of the time series at time $t$, and they represent the past state; $x_{t+1}$ and $y_{t+1}$ represent the future state. In addition, $x_{t}^{(k)} = [x_{t}, $ $x_{t-1}, ..., x_{t-k+1}]$ and $y_{t}^{(l)} = [y_{t}, y_{t-1}, ..., y_{t-l+1}]$.
	
	\begin{equation} \label{eq3}
		\begin{aligned}
			TE_{y \rightarrow x}
			= & \sum p\left(x_{t+1}, {x}_{t}^{({k})}, {y}_{t}^{(l)}\right) \log _{2} \frac{p\left(x_{t+1} \mid {x}_{t}^{(k)}, {y}_{t}^{(l)}\right)}{p\left(x_{t+1} \mid {x}_{t}^{(k)}\right)} \\
			= & \sum p\left(x_{t+1}, {x}_{t}^{(k)}, {y}_{t}^{(l)}\right) \log _{2} p\left(x_{t+1} \mid {x}_{t}^{(k)}, {y}_{t}^{(l)}\right) \\
			& -\sum p\left(x_{t+1}, {x}_{t}^{({k})}\right) \log _{2} p\left(x_{t+1} \mid {x}_{t}^{(k)}\right) \\
			= & H\left({X}_{t+1} \mid {X}_{t}\right)-H\left({X}_{t+1} \mid {X}_{t}, {Y}_{t}\right),
		\end{aligned}
	\end{equation}
	
	In Eq.\eqref{eq3}, $H\left(X_{t+1} \mid X_{t}\right)-H\left(X_{t+1} \mid X_{t}, Y_{t}\right)$ is the conditional entropy, which means that the transfer entropy from $Y$ to $X$ represents the reduction of uncertainty in the value of $X$ when the past value of $X$ is known. Here, $X_{t}$ and $Y_{t}$ represent the past state, whereas $X_{t+1}$ and $Y_{t+1}$ represent the future state. Conditional entropy $H(X \vert Y)$ represents the amount of information in $X$ given the known condition of $Y$. Moreover, $p(x_{t+1}, {x}_{t}^{({k})}, {y}_{t}^{(l)})$ represents the joint probability distribution of future state $x_{t+1}$ given past states ${x}_{t}^{({k})}$ and ${y}_{t}^{(l)}$; $p(x_{t+1} \mid {x}_{t}^{(k)}, {y}_{t}^{(l)})$ represents the conditional probability distribution of future state $x_{t+1}$ given past states ${x}_{t}^{(k)}$ and ${y}_{t}^{(l)}$; $p(x_{t+1} \mid {x}_{t}^{(k)})$ represents the conditional probability distribution of future state $x_{t+1}$ given past state ${x}_{t}^{(k)}$; $H\left({X}_{t+1} \mid {X}_{t}\right)$ represents the conditional entropy of future state ${X}_{t+1}$ given past state ${X}_{t}$; $H\left({X}_{t+1} \mid {X}_{t}, {Y}_{t}\right)$ represents the conditional entropy of future state ${X}_{t+1}$ given past states ${X}_{t}$ and ${Y}_{t}$.
	
	For two dimensions $X$ and $Y$, the conditional entropy is defined as follows, where $x$ denotes all possible values in dimension $X$ and $y$ denotes all possible values in dimension $Y$.

	\begin{equation} \label{eq2}
			H(X \vert Y) = -\sum \sum p(x,y)\log_{2} (p(x \vert y))
	\end{equation}
	
	When the transfer entropy $TE_{X\rightarrow Y}$ is larger than the transfer entropy $TE_{Y\rightarrow X}$, a causal correlation between the two variables is established. The causal correlation between $X$ and $Y$ can be further defined as follows. 
	
	\begin{equation} \label{eq4}
		C_{X,Y} = TE_{X\rightarrow Y} - TE_{Y\rightarrow X}
	\end{equation}
	If $C_{X,Y}$ is greater than 0, we say that $X$ affects $Y$.
	
\begin{figure}[!tb]
		\centering
		\includegraphics[width=0.7\textwidth]{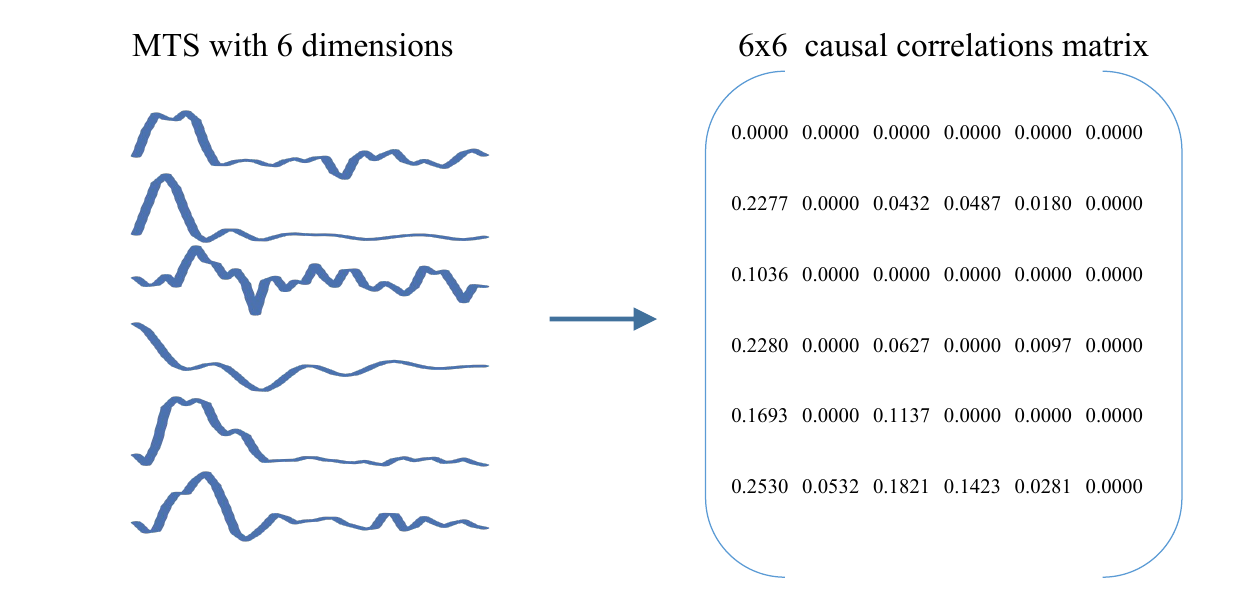}
		\caption{Example of a causal correlation matrix.}
		\label{fig:fig3}
	\end{figure}

	Finally, the causal correlations $TE$ of the MTS are constructed in a matrix $M$. The dimensions of $M$ are $n \times n$ (where $n$ is the number of dimensions in $T$). The value $M_{i,j}$ in the $i$-th row and $j$-th column of $M$ can be calculated as follows, where $T_{i}$ represents the $i$-th dimension of MTS $T$, $T_{j}$ represents the $j$-th dimension of $T$, and $c$ is the threshold value used to determine whether the causal correlation is significant. An example of a causal correlation matrix is shown in Fig.\ref{fig:fig3}, in which a $6 \times 6$ causal correlations matrix has been obtained for an MTS with six dimensions using this procedure.
	
	\begin{equation} \label{eq5}
		M_{i,j}=\left\{
		\begin{array}{rcl}
			C_{T_{i},T_{j} },& &C_{T_{i},T_{j}  }>c\\
			0,& & otherwise
		\end{array}
		\right.
	\end{equation}
	
	Transfer entropy is a method used to quantify the causal relationships between MTSs. It is based on the concept of information theory and measures the "transfer" or "propagation" of information from one time series to another. Transfer entropy helps identify causal dependencies between time series, i.e., how the values of one time series influence the values of another. By computing transfer entropy, we can quantify the strength of causal relationships between time series. A higher value of transfer entropy indicates a stronger influence of X on Y, suggesting a stronger causal relationship. Conversely, a value close to zero indicates a weaker influence of X on Y, implying a lack of obvious causality.

	Transfer entropy can be used to compute a causal relationship matrix between MTSs and plays an important role in the GNN. Transfer entropy can be used to construct causal relationship graphs by computing the causal relationships between time series. In the GNN, these causal relationship graphs can be represented as a graph structure, where nodes represent time series and edges represent causal relationships between time series. For instance, in Section \ref{sec3.4}, the causal correlations matrix has the graph structure of A graph isomorphism network (GIN). Such causal relationship graphs capture complex dependencies between time series and provide a foundation for subsequent analysis and prediction tasks. The causal relationship matrix obtained from transfer entropy calculation can serve as one of the inputs to the GNN. In the GNN, node features are typically used to represent attributes of time series, whereas edges represent relationships between time series. The causal relationship matrix can be used to define the adjacency matrix of the graph, explicitly representing the causal relationships between time series. Thus, a GNN can learn the causal propagation patterns between time series, thereby improving the accuracy of prediction and analysis.

	In summary, the application of transfer entropy in a GNN helps model and capture causal relationships between MTSs, enhancing the predictive and analytical capabilities of the models. By combining the strengths of transfer entropy and the GNN, we can better understand the dynamic characteristics and interactions of time series data, thereby advancing research and applications in related fields.
	
	\subsection{Local correlation extraction}
	\label{sec3.3}
	
	CaLoNet extracts local correlations as shown in Fig.\ref{fig:fig2}. The following main processes are used to extract local correlations:
	\begin{itemize}
		\item \textbf{EMBED}. First, MTS is partitioned. We define the initial embedding of an MTS as $T$ (with $D$ dimensions, where each dimension has a length of $L$). CaLoNet uses four non-overlapping neighboring timestamps to obtain $\frac{L}{4}$ time chunks. Each time chunk is then flattened and projected into a $4C$-dimensional embedding. Finally, we obtain an $\frac{L}{4} \times 4D$-dimensional embedding, denoted as $E$, where $X_{embed}$ has a dimension of $4D$ and the length of each dimension is $\frac{L}{4}$.
		\item \textbf{CBAM}. The convolutional block attention module (CBAM) is a network used for feature refinement. We describe CBAM in Section \ref{sec:cbam}.
		\item \textbf{SSA}. The SSA layer is a network used for extracting local correlations. We describe SSA in Section \ref{sec:ssa}.
		\item \textbf{LN}. The layer normalization(LN) layer normalizes the hidden layers in the network to a standard normal distribution to speed up training and accelerate convergence.
		\item \textbf{MLP}. The MLP layer functions as a fully connected layer.
		\item \textbf{SHIFT}. The above process, except for the embedding layer, is repeated twice in the local correlation layer. However, a shift layer is added in the second pass \cite{liu2021swin}, which moves the temporal blocks within the window. This solves the problem of global features being restricted to the local window partition. Each patch can then interact with other patches in a new window.
	\end{itemize}
	
	\begin{figure}[!tb]
		\centering
		\includegraphics[width=0.95\textwidth]{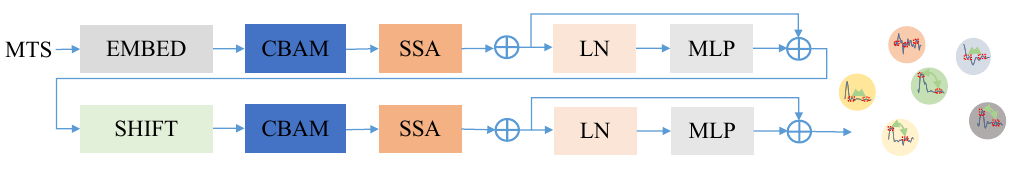}
		\caption{Local correlation network.}
		\label{fig:fig4}
	\end{figure}
	
	\subsubsection{CBAM layer}
	\label{sec:cbam}
	
	Effective feature capture plays a crucial role in the MTS classification task. The CBAM layer \cite{woo2018cbam} shares similarities with the SE block \cite{karim2019multivariate, hu2018squeeze}, as both leverage attention mechanisms for feature refinement. The attention mechanism facilitates the refinement of features, whereas the self-attention mechanism establishes connections between different positions of a time series to derive relationships at specific positions. The CBAM layer generates attention maps along two independent dimensions: channel and spatial. These attention maps are then multiplied by the input feature map to adaptively refine features.
	
\begin{figure}[!tb]
		\centering
		\includegraphics[width=0.7\textwidth]{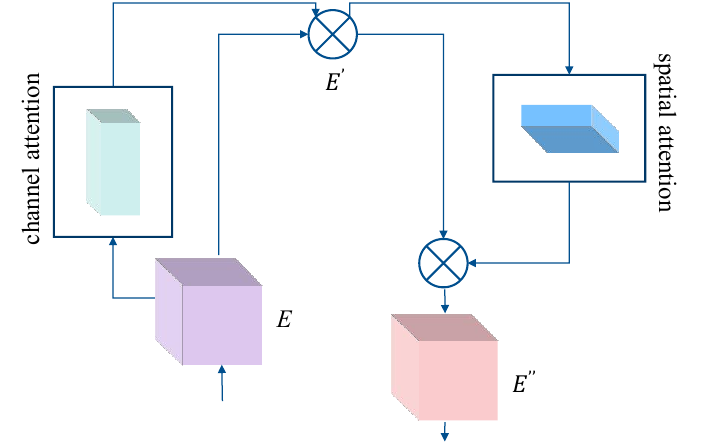}
		\caption{CBAM block.}
		\label{fig:fig5}
	\end{figure}

	First, MTS embedding is performed through the CBAM layer, which is a module designed to exploit spatial and channel attention mechanisms to focus on more discriminative features. As shown in Fig.\ref{fig:fig5}, channel attention is used to obtain $E^{'}$ by employing average pooling and maximum pooling operations, ultimately aggregating the spatial information of the feature map. The channel attention is calculated as follows, where $W_{1}$ and $W_{0}$ represent the shared parameters of the $MLP$ and $E$ denotes the features from the EMBED layer.
	\begin{equation} \label{eq6}
		\begin{split}
			E^{'} &=\mathit{sigmoid} (\mathit{MLP} (\mathit{AvgPool} (E))+ \mathit{MLP} (\mathit{MaxPool} (E))) \\
			&=\mathit{sigmoid} (W_{1}(W_{0}({E }_{avg}^{c}))+W_{1}(W_{0}({E}_{max}^{c})))
		\end{split}
	\end{equation}
	
	The spatial attention mechanism is used to obtain the spatial attention $E^{''}$. $E^{'}{_{avg}^{s}}$  and $E^{'}{_{max}^{s}}$ represent the average pooling operation and the maximum pooling operation applied along the channel axis. Then, the spatial attention feature map is generated by the one-dimensional CNN, it is computed as follows, where $\mathit{f^{a}}$ denotes a one-dimensional convolution operation with filter size $a$, and ${E^{'}}$ are features from the channel attention layer.
	\begin{equation} \label{eq7}
		\begin{split}
			E^{''}&=\mathit{sigmoid} (\mathit{f^{a}} ([\mathit{AvgPool} (E^{'} ) ;\mathit{MaxPool} (E^{'} )])) \\
			&=\mathit{sigmoid} (\mathit{f^{a}} ([{E^{'}}_{avg}^{s};{E^{'}}  _{max}^{s}]))		
		\end{split}
	\end{equation}
	
	\subsubsection{SSA layer} 
	\label{sec:ssa}
	The SSA layer, which is necessary for effective local correlation extraction, is used by CaLoNet to efficiently extract local correlations, resulting in significant time savings. For instance, \cite{liu2021swin, vaswani2017attention} explored both long-term and short-term dependencies in the data, thereby enhancing relationship extraction. The incorporation of the CBAM into the self-attention structure theoretically enables the extraction of more time-sensitive features and higher sensitivity to global changes in non-periodic data than fully connected networks.
	
\begin{figure}[!tb]
		\centering
		\includegraphics[width=0.7\textwidth]{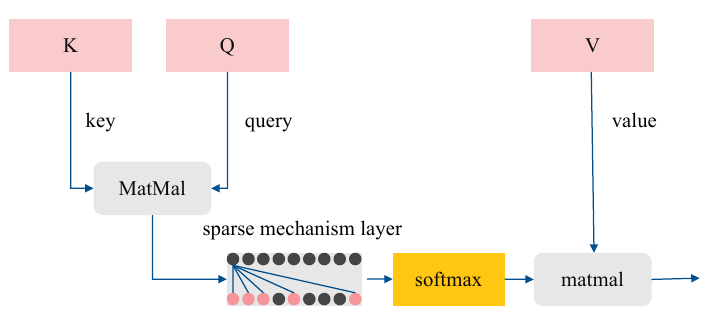}
		\caption{SSA mechanism.}
		\label{fig:fig6}
	\end{figure}

	As depicted in Fig.~\ref{fig:fig6}, the dot product score is calculated by multiplying $Q$ and $K$ to obtain the computed weight coefficient and compute the similarity first.
	
\begin{figure}[!tb]
		\centering
		\includegraphics[width=0.7\textwidth]{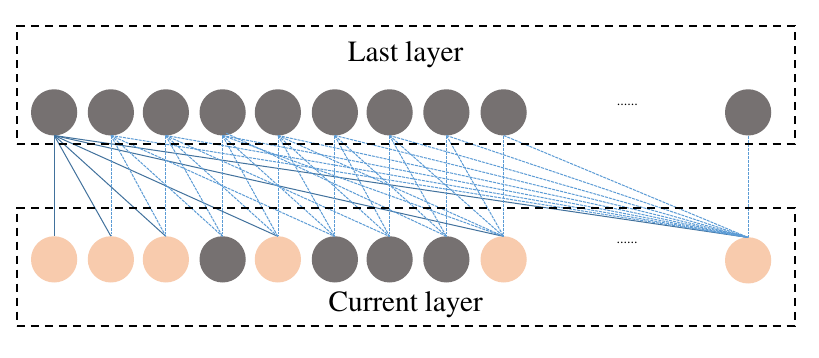}
		\caption{Principle of the SSA mechanism. The SSA is computed by only allowing each cell of the previous layer to pay attention to the cell of the current layer in exponential steps. The dark circles in the current layer do not participate in the dot product operation, whereas the bright circles do participate.}
		\label{fig:fig7}
	\end{figure}

	Next, the SSA mechanism is used. The heads of each layer may focus on different types of features, as described by \cite{li2019enhancing}, when stacking Transformers. For example, for a time series generated monthly, the heads of each layer will focus on the features of a particular week. Therefore, we restrict the cells to two adjacent layers. That is, we only allow each layer of cells to participate in the dot product operation for those cells with previous exponential steps. Thus, the SSA layer only needs to compute the $O(\log L)$ ($L$ is the time series length) dot product in each layer, as shown in Fig.~\ref{fig:fig7}.
	
	Furthermore, to handle the SSA layer's memory bottleneck problem, the space complexity of the normal transformer grows quadratically with the increase in time series length $L$. The time complexity of $O(L^{2})$ results in huge memory consumption \cite{zhou2021informer}. Therefore, to compute the self-attention dot product more efficiently, we adopt a sparse strategy \cite{li2019enhancing} to design the attention layer. We introduce a sparse bias matrix $M$ in the self-attention model with a log-sparse mask, as shown in Eq.\eqref{eq9} and Fig.\ref{fig:fig7}. In this approach, only the $O(\log L)$ dot product of each cell in each layer must be computed, thus reducing the computational complexity from $O(L^{2})$ to $O(L \log L)$.
	
	Then, the softmax activation function is applied to the score obtained through the incorporated SSA mechanism. Finally, the softmax dot product $V$ is calculated to obtain the weighted score for each input vector.
	
	The query, key, and value matrices are matrices obtained by extracting features from the CBAM layer. Specifically, embedding $E$ $(E_{embed})$ is represented using matrices $Q$, $K$, and $V$, as shown in Eq.\eqref{eq8}. The dot product for the self-attention mechanism is then calculated using Eq.\eqref{eq9}, as shown in Fig.\ref{fig:fig6}. In Fig.\ref{fig:fig6}, first, the dot product score is calculated by multiplying $Q$ and $V$ to obtain the computed weight coefficients and similarity. Then, the softmax activation function is applied to the score through the incorporated SSA mechanism. Finally, the softmax dot product $V$ is used to obtain the weighted score for each input vector.
	\begin{equation} \label{eq8}
		\begin{split}
			Q=\mathit{CBAM}_{Q} (E_{embed}  )\\
			K=\mathit{CBAM}_{K} (E_{embed}  )\\
			V=\mathit{CBAM}_{V} (E_{embed}  )		\\
		\end{split}
	\end{equation}
	\begin{equation} \label{eq9}
		\begin{split}
			F =\mathit{SoftMax} (\frac{QK^{T}  }{\sqrt{d_{k} } }+B )V
		\end{split}
	\end{equation}
	
	In Eq.\eqref{eq8}, $CBAM$ represents the function of the CBAM layer, which is used to obtain the query, key, and value matrices. In Eq.\eqref{eq9}, $SoftMax$	  is a function commonly used to compute a probability distribution that takes a vector as input and converts it to a vector of probability distributions. It is often used in classification tasks to represent the probability of different categories. In addition, $\sqrt{d_{k}}$ is used for score normalization, $K^{T}$ represents the transpose of $K$, and $B$ is a sparse bias matrix based on Fig.~\ref{fig:fig7}.
	
	\subsection{Node embedding}\label{sec3.4}
	\begin{figure}[!tb]
		\centering
		\includegraphics[width=0.8\textwidth]{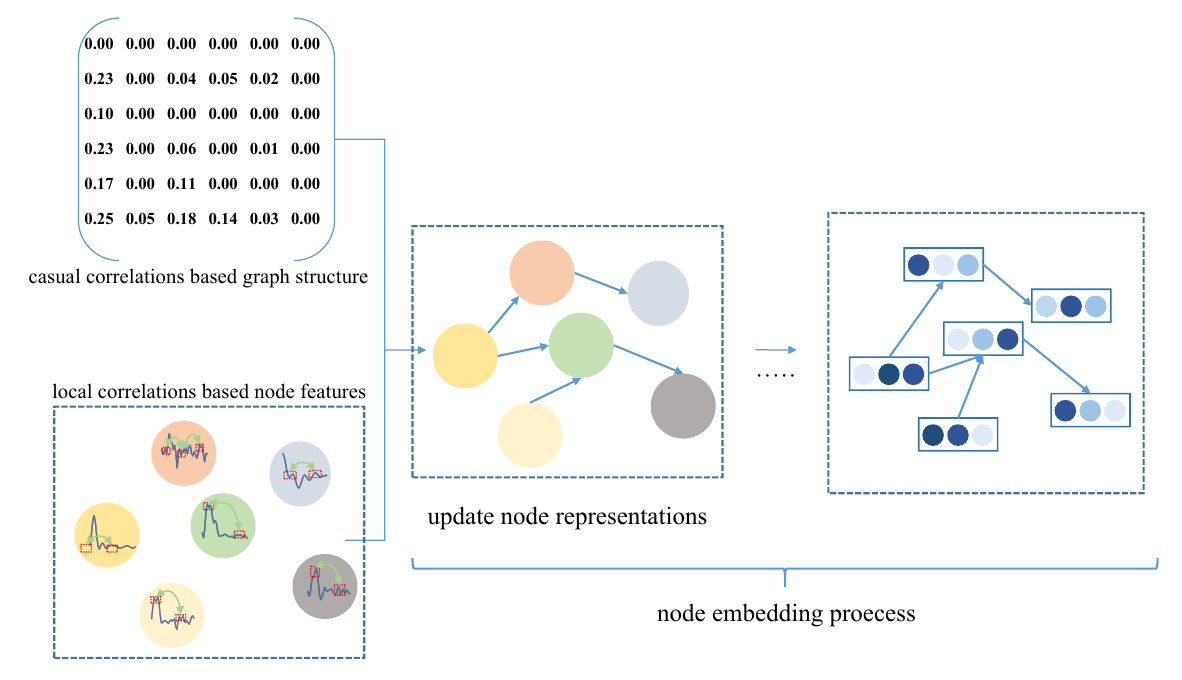}
		\caption{Node embedding process.}
		\label{nodeembed}
	\end{figure}

	After obtaining the graph structure as described in Section~\ref{sec3.2} and the node features as described in Section~\ref{sec3.3}, the method updates the node features to obtain the final node representations (right panel in Fig.~\ref{nodeembed}). Using the different working mechanisms in the two approaches, namely local correlations and spatial correlations, we explore the possibility of connecting these two approaches to jointly model MTS learning representations. However, existing deep-learning methods cannot explicitly represent MTS spatial correlations.
	The input MTS is converted into a feature matrix $H \in F_{n \times d}$, where $d$ is the calculated number of features, as introduced in Section~\ref{sec3.2}. Matrix $H$ can be viewed as a graph feature matrix with $n$ nodes, which correspond to the six variables of the MTS. The adjacency of nodes in the graph structure is determined by causal matrix $T$. For this graph structure, a GNN can be directly applied to the node embeddings. Inspired by the GIN model \cite{xu2018powerful}, the following propagation mechanism is used to compute the forward-passing update for a node represented by $v$. GIN updates the node representations as shown in Eq.~\eqref{eq10}.

	After processing with the GIN model, the feature vectors $h_v^{(k)}$ for each node are computed. The GIN model updates node features by iteratively applying MLP operations $MLP^k$. An MLP is a multi-layered neural network that uses nonlinear transformations to process node features. For each node $v$, its feature vector is updated at each layer $k$ to capture information about the node at that specific layer.

	When updating node features, the GIN model incorporates a tunable parameter $\epsilon^k$ specific to each layer $k$. This parameter adaptively controls the influence of the previous layer's features $h_v^{(k-1)}$ on the current layer's features $h_v^{(k)}$. By adjusting the value of $\epsilon^k$, the model can balance the propagation and aggregation of features across different layers.

	During the computation of node features, the GIN model also leverages information from the node's neighbors. $N(v)$ represents the set of neighbors of node $v$, which includes neighboring nodes $u$. For each node $v$, the GIN model aggregates the feature vectors $h_u^{(k-1)}$ of its neighboring nodes by summing them. This aggregation of neighbor information helps the synthesis of node features and contextual modeling.
	\begin{equation} \label{eq10}
		h_{v}^{(k)} =MLP^{k} ((1+\epsilon^{k} )\cdot h_{v}^{k-1} + {\textstyle \sum_{\mu \in N(v)}^{h_{u}^{(k-1)} }}  )
	\end{equation}
	
	In Eq.~\eqref{eq10}, 
	$h_v^{(k)}$ represents the feature vector of node $v$ at layer $k$---this vector captures the information about the node at that particular layer;
	$MLP^k$ denotes the MLP applied to the features at layer $k$, where the MLP is a feedforward neural network with multiple layers of neurons that transform the input features;
	$\epsilon^k$ is an updatable parameter specific to layer $k$ that allows the model to adaptively control the influence of the previous layer's features $h_v^{(k-1)}$ on the current layer's features $h_v^{(k)}$;
	$N(v)$ represents the neighborhood of node $v$, which consists of neighboring nodes $u$;
	$h_u^{(k-1)}$ represents the feature vector of the neighboring node $u$ at layer $k-1$. The sum is taken over all neighboring nodes to aggregate their features.
	
	\subsection{Classification}\label{sec3.5}
	
	Finally, CaLoNet classifies the final features $X$ using an MLP. In the MLP, $X$ is converted to predicted class labels using a fully connected layer. We trained the MLP using the loss function shown in Eq.\eqref{eq11}, where $N$ represents the number of samples in the training set, $M$ is the number of labels, $y_{i,j}$ denotes the true label of the i-th sample belonging to the j-th class, and $\hat{y}_{i,j}$ denotes the probability that the model predicts that the i-th sample belongs to the j-th class. The loss function is in the form of cross entropy, which measures the difference between the true label y and the predicted probability $\hat{y}$.
	
	\begin{equation} \label{eq11}
			\ell(X) = -\frac{1}{N} \sum_{i=1}^{N} \sum_{j=1}^{M} y_{i,j} \log p(\hat{y}_{i,j})
	\end{equation}
	
	\section{Experiment}
	\label{sec:exp}
	
	\subsection{Experimental setting}
	\subsubsection{Datasets}
	Twenty-one multivariate datasets from the UEA archive\footnote{ Datasets are available at http://timeseriesclassification.com} were used for our experiments. Table~\ref{tab1} lists their detailed information.
	\begin{table}[!tb]
		\centering
		\caption{Experimental datasets.}
		\label{tab1}
		\footnotesize
		\begin{tabular}{llllllll}
			\hline
			Dataset & Abbrevation & Type & Train & Test & Dimensions & Length & Classes \\ \hline
			ArticularyWordRecognition & AWR & Motion & 275 & 300 & 9 & 144 & 25 \\
			AtrialFibrillation & AF & ECG & 15 & 15 & 2 & 640 & 3 \\
			BasicMotions & BM & HAR & 40 & 40 & 6 & 100 & 4 \\
			CharacterTrajectories & CT & Motion & 1422 & 1436 & 3 & 182 & 20 \\
			Cricket & CR & HAR & 108 & 72 & 6 & 1197 & 12 \\
			EthanolConcentration & EC & HAR & 261 & 263 & 3 & 1751 & 4 \\
			FaceDetection & FD & EEG/MEG & 5890 & 3524 & 144 & 62 & 2 \\
			HandMovementDirection & HMD & EEG/MEG & 160 & 74 & 10 & 400 & 4 \\
			Heartbeat & HB & AS & 204 & 205 & 61 & 405 & 2 \\
			JapaneseVowels & JV & AS & 270 & 370 & 12 & 29 & 9 \\
			Libras & LIB & HAR & 180 & 180 & 2 & 45 & 15 \\
			LSST & LSST & Other & 2459 & 2466 & 6 & 36 & 14 \\
			MotorImagery & MI & EEG/MEG & 278 & 100 & 64 & 3000 & 2 \\
			NATOPS & NA & HAR & 180 & 180 & 24 & 51 & 6 \\
			PEMS-SF & PEMS & Other & 267 & 173 & 963 & 144 & 7 \\
			PenDigits & PD & Motion & 7494 & 3498 & 2 & 8 & 10 \\
			SelfRegulationSCP1 & SR1 & EEG/MEG & 268 & 293 & 6 & 896 & 2 \\
			SelfRegulationSCP2 & SR2 & EEG/MEG & 200 & 180 & 7 & 1152 & 2 \\
			SpokenArabicDigits & SAD & AS & 6599 & 2199 & 13 & 93 & 10 \\
			StandWalkJump & SWJ & ECG & 12 & 15 & 4 & 2500 & 3 \\
			UWaveGestureLibrary & UW & HAR & 120 & 320 & 3 & 315 & 8   \\ \hline   
		\end{tabular}
	\end{table}
	
	\subsubsection{Experimental running environment}
	Our experiments were conducted on a Windows computer using Python 3.9 and PyTorch 1.10. Minimization was performed using a cross-entropy loss function and the ADAM optimizer. The results were obtained after 50 iterations for all datasets.
	
	\subsubsection{Reproducibility}
	For reproducibility, we released our code and parameters on GitHub.\footnote{ https://github.com/dumingsen/CaLoNet} The experimental results can be independently replicated.
	
	\subsection{Comparison with baselines}
	CaLoNet was compared with 16 baselines: SMATE \cite{zuo2021smate},	MLSTM-FCN \cite{karim2019multivariate},	 WEASEL+MUSE \cite{schafer2017multivariate},	DA-Net \cite{chen2022net},	MR-PETSC \cite{feremans2022petsc}, 	ED-1NN \cite{chen2013dtw},	DTW-1NN-I \cite{chen2013dtw},	DTW-1NN-D \cite{chen2013dtw},	ED-1NN(norm) \cite{chen2013dtw},	DTW-1NN-I(norm) \cite{chen2013dtw},	11: DTW-1NN-D(norm) \cite{chen2013dtw},	 RT(100\%) \cite{Bahri2022ShapeletbasedTA},	RT(20\%) \cite{Bahri2022ShapeletbasedTA},	MF-Net \cite{Du2023MultifeatureBN},	TapNet \cite{zhang2020tapnet},	and MTSC\_FF \cite{Du2024MultivariateTS}.  
	
	The accuracy results of these methods are listed in Table~\ref{tab2}. The results for the comparison methods were obtained from SMATE \cite{zuo2021smate}, DA-Net \cite{chen2022net}, MR-PETSC \cite{feremans2022petsc}, MF-Net \cite{Du2023MultifeatureBN}, MTSC\_FF \cite{Du2024MultivariateTS}, and RT \cite{Bahri2022ShapeletbasedTA}. The highest accuracy for each dataset is indicated in bold. In Table~\ref{tab2}, "AVG" represents the average accuracy across the 21 datasets, and "Win" indicates the number of datasets on which the method achieved the best performance. Table~\ref{tab2} reveals that CaLoNet achieves the highest average accuracy among these methods. Moreover, CaLoNet obtains the best accuracy on five datasets. These experimental results demonstrate the superior performance of CaLoNet.
	
\begin{figure}[!tb]
		\centering
		\includegraphics[width=0.8\textwidth]{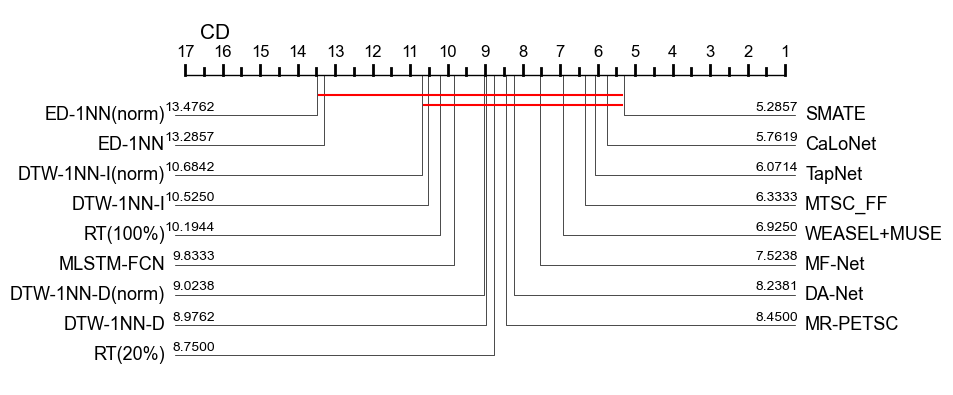}
		\caption{Critical difference diagram for the proposed method and baseline methods.}
		\label{fig:fig8}
	\end{figure}

	To evaluate the differences between CaLoNet and the other baseline methods, we present a critical difference diagram in Fig.\ref{fig:fig8} based on the accuracies in Table~\ref{tab2}. In this diagram, CaLoNet has the second smallest ranking. This critical difference diagram further illustrates the superiority of our method.

	In the critical difference plot, CaLoNet ranks second to SMATE. Hence, we used the Wilcoxon rank-sum test \cite{Wilcoxon1945IndividualCB} to validate the performance difference between CaLoNet and SMATE. According to the results, the p-value (0.9709) is greater than the commonly used significance level (e.g., 0.05), indicating that we do not have enough evidence to reject the null hypothesis. The statistic (-0.0364) is close to zero, suggesting that the difference between the two samples is very small. From the given statistic and p-value, we can infer that the two samples are similar or not significantly different. This means that we lack sufficient evidence to suggest a significant difference between the two samples. 
	
	SMATE extracts spatio--temporal dynamic features for each dimension subsequence, emphasizing the significance of temporal dependency and spatial interactions when constructing reliable embeddings for MTS. However, SMATE is unable to explicitly obtain graph structures for the spatial features, limiting the interpretation of spatial correlations through visualization methods. Our approach achieves accuracy similar to that of SMATE, but we further leverage causal relationships to capture spatial correlations and obtain interpretable graph structures.
	
		\begin{sidewaystable*}
			\centering
			\caption{Comparison of the accuracy results of CaLoNet and the baseline methods.}
			\label{tab2}
			\scalebox{0.8}{		
				\begin{tabular}{llllllllllllllllll}
					\hline
					Dataset & 1 & 2 & 3 & 4 & 5 & 6 & 7 & 8 & 9 & 10 & 11 & 12 & 13 & 14 & 15 & 16 & 17 \\ 	\hline
					AWR & 0.993 & 0.973 & 0.990 & 0.980 & \textbf{0.997} & 0.970 & 0.980 & 0.987 & 0.970 & 0.980 & 0.987 & 0.993 & 0.990 & 0.983 & 0.987 & 0.983 & 0.983 \\
					AF & 0.133 & 0.267 & 0.333 & 0.467 & 0.400 & 0.267 & 0.267 & 0.200 & 0.267 & 0.267 & 0.220 & 0.200 & 0.266 & 0.466 & 0.333 & \textbf{0.533} & 0.414 \\
					BM & \textbf{1.000} & 0.950 & \textbf{1.000} & 0.925 & \textbf{1.000} & 0.675 & \textbf{1.000} & 0.975 & 0.676 & \textbf{1.000} & 0.975 & \textbf{1.000} & \textbf{1.000} & 0.950 & \textbf{1.000} & 0.950 & 0.975 \\
					CT & 0.984 & 0.985 & 0.990 & \textbf{0.998} & 0.941 & 0.964 & 0.969 & 0.990 & 0.964 & 0.969 & 0.989 & N/A & N/A & 0.958 & 0.997 & 0.986 & 0.987 \\
					CR & 0.986 & 0.917 & \textbf{1.000} & 0.861 & \textbf{1.000} & 0.944 & 0.986 & \textbf{1.000} & 0.944 & 0.986 & \textbf{1.000} & \textbf{1.000} & \textbf{1.000} & 0.944 & 0.958 & 0.986 & 0.986 \\
					EC & 0.399 & 0.373 & 0.430 & \textbf{0.874} & 0.555 & 0.293 & 0.304 & 0.323 & 0.293 & N/A & 0.323 & 0.395 & 0.414 & 0.250 & 0.323 & 0.292 & 0.395 \\
					FD & 0.647 & 0.545 & 0.545 & 0.648 & 0.574 & 0.519 & 0.513 & 0.529 & 0.519 & 0.500 & 0.529 & 0.507 & 0.526 & 0.664 & 0.556 & \textbf{0.670} & 0.648 \\
					HMD & \textbf{0.554} & 0.365 & 0.365 & 0.365 & 0.338 & 0.279 & 0.306 & 0.231 & 0.278 & 0.306 & 0.231 & 0.337 & 0.405 & 0.500 & 0.378 & 0.540 & 0.527 \\
					HB & 0.741 & 0.663 & 0.730 & 0.624 & 0.702 & 0.620 & 0.659 & 0.717 & 0.619 & 0.658 & 0.717 & 0.726 & 0.731 & 0.682 & \textbf{0.751} & 0.692 & 0.682 \\
					JV & 0.965 & 0.976 & 0.973 & 0.938 & N/A & 0.924 & 0.959 & 0.949 & 0.924 & 0.959 & 0.949 & N/A & N/A & 0.970 & 0.965 & 0.978 & \textbf{0.981} \\
					LIB & 0.849 & 0.856 & 0.878 & 0.800 & 0.845 & 0.833 & \textbf{0.894} & 0.872 & 0.833 & \textbf{0.894} & 0.870 & 0.855 & 0.866 & 0.850 & 0.850 & 0.861 & 0.850 \\
					LSST & 0.582 & 0.373 & \textbf{0.590} & 0.560 & 0.560 & 0.456 & 0.575 & 0.551 & 0.456 & 0.575 & 0.551 & 0.336 & 0.366 & 0.468 & 0.568 & 0.478 & 0.560 \\
					MI & \textbf{0.590} & 0.510 & 0.510 & 0.500 & 0.490 & 0.390 & N/A & 0.500 & 0.510 & N/A & 0.500 & 0.500 & 0.510 & 0.540 & \textbf{0.590} & 0.550 & \textbf{0.590} \\
					NA & 0.922 & 0.889 & 0.870 & 0.878 & 0.917 & 0.860 & 0.850 & 0.883 & 0.850 & 0.850 & 0.883 & 0.877 & 0.872 & 0.927 & \textbf{0.939} & 0.888 & 0.916 \\
					PEMS & 0.803 & 0.699 & N/A & 0.867 & 0.861 & 0.705 & 0.734 & 0.711 & 0.705 & 0.734 & 0.711 & 0.364 & 0.358 & \textbf{0.884} & 0.751 & \textbf{0.884} & 0.878 \\
					PD & 0.980 & 0.978 & 0.948 & 0.980 & 0.905 & 0.973 & 0.939 & 0.977 & 0.973 & 0.939 & 0.977 & 0.620 & 0.620 & \textbf{0.983} & 0.980 & 0.979 & \textbf{0.983} \\
					SR1 & 0.887 & 0.874 & 0.710 & 0.924 & 0.788 & 0.771 & 0.765 & 0.775 & 0.771 & 0.765 & 0.775 & 0.856 & 0.863 & 0.911 & 0.739 & \textbf{0.928} & 0.907 \\
					SR2 & \textbf{0.567} & 0.472 & 0.460 & 0.561 & 0.533 & 0.483 & 0.533 & 0.539 & 0.483 & 0.533 & 0.539 & 0.477 & 0.538 & 0.533 & 0.550 & 0.510 & 0.505 \\
					SAD & 0.979 & \textbf{0.990} & 0.982 & 0.980 & 0.960 & 0.967 & 0.960 & 0.963 & 0.967 & 0.959 & 0.963 & N/A & N/A & \textbf{0.990} & 0.983 & \textbf{0.990} & \textbf{0.990} \\
					SWJ & \textbf{0.533} & 0.067 & 0.333 & 0.400 & 0.400 & 0.200 & 0.333 & 0.200 & 0.200 & 0.333 & 0.200 & 0.400 & 0.333 & 0.400 & 0.400 & 0.466 & \textbf{0.533} \\
					UW & 0.897 & 0.891 & \textbf{0.916} & 0.833 & 0.800 & 0.881 & 0.868 & 0.903 & 0.810 & 0.868 & 0.903 & 0.875 & 0.875 & 0.862 & 0.894 & 0.875 & 0.881 \\
					AVG & 0.761 & 0.695 & 0.727 & 0.760 & 0.728 & 0.665 & 0.719 & 0.703 & 0.667 & 0.740 & 0.704 & 0.628 & 0.640 & 0.748 & 0.737 & 0.762 & \textbf{0.770} \\
					Win & 5 & 1 & 4 & 2 & 3 & 0 & 2 & 1 & 0 & 2 & 1 & 2 & 2 & 3& 4 & \textbf{5} & \textbf{5} \\ \hline
				\end{tabular}
			}
			\begin{tablenotes} 
				\item The classifiers (1-17) compared in the table are as follows. 1: SMATE \cite{zuo2021smate},	2: MLSTM-FCN \cite{karim2019multivariate},	3: WEASEL+MUSE \cite{schafer2017multivariate},	4: DA-Net \cite{chen2022net},	5: MR-PETSC \cite{feremans2022petsc},	6: ED-1NN \cite{chen2013dtw},	7: DTW-1NN-I \cite{chen2013dtw},	8: DTW-1NN-D \cite{chen2013dtw},	9: ED-1NN(norm) \cite{chen2013dtw},	10: DTW-1NN-I(norm) \cite{chen2013dtw},	11: DTW-1NN-D(norm) \cite{chen2013dtw},	12: RT(100\%) \cite{Bahri2022ShapeletbasedTA},	13: RT(20\%) \cite{Bahri2022ShapeletbasedTA},	14: MF-Net \cite{Du2023MultifeatureBN},	15: TapNet \cite{zhang2020tapnet},	16: MTSC\_FF \cite{Du2024MultivariateTS},	17: CaLoNet (our method). 
				
			\end{tablenotes} 
		\end{sidewaystable*}
	
	\subsection{Ablation study}
	
	\subsubsection{Comparison of dependent components}

\begin{figure}[!tb]
		\centering
		\subfloat[CaLoNet vs. Only LCP]{
			\centering
			\includegraphics[width=0.45\textwidth]{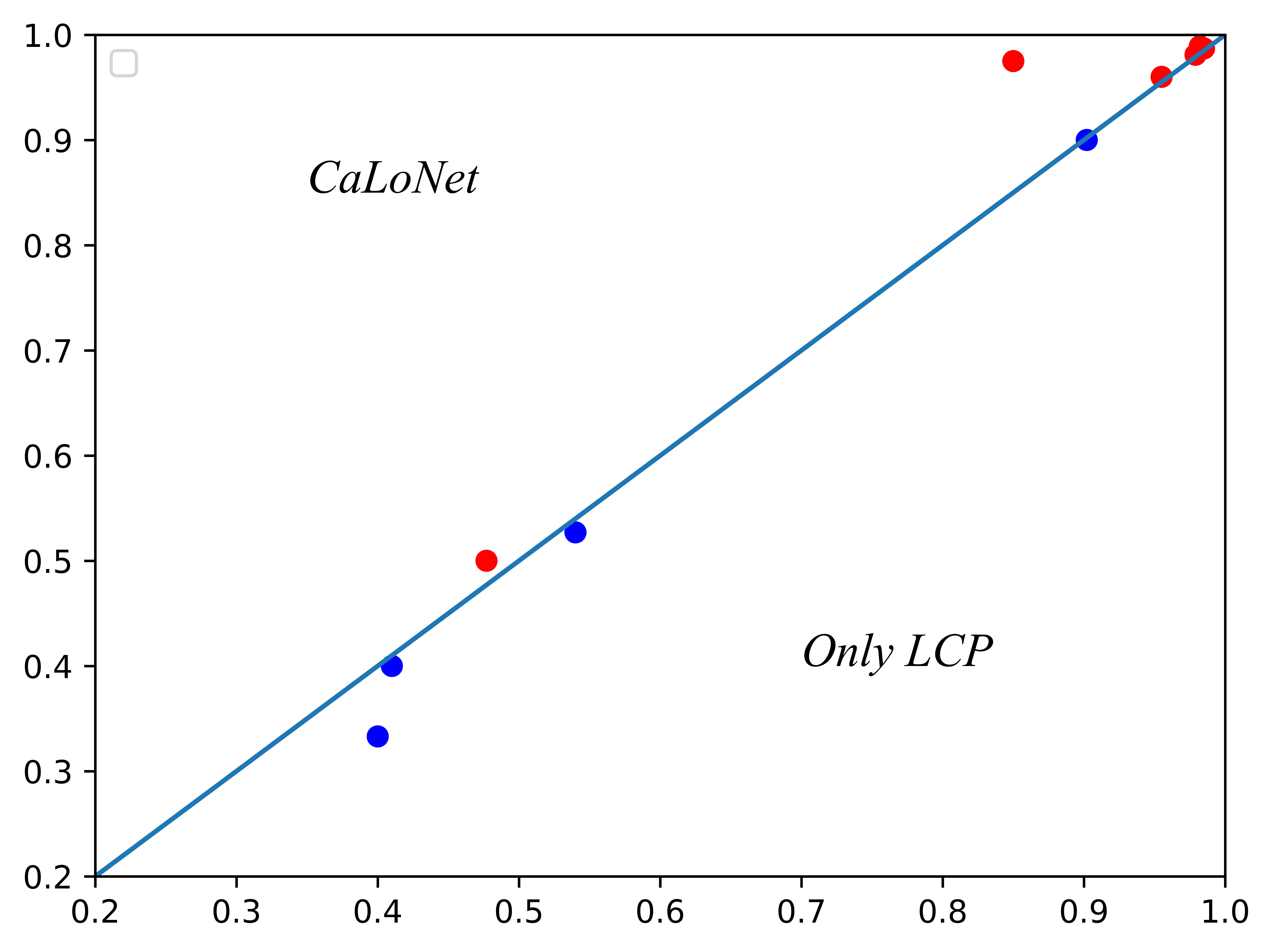}
			\label{fig:fig9a}
		}
		\subfloat[CaLoNet vs. Only CCP]{
			\centering
			\includegraphics[width=0.45\textwidth]{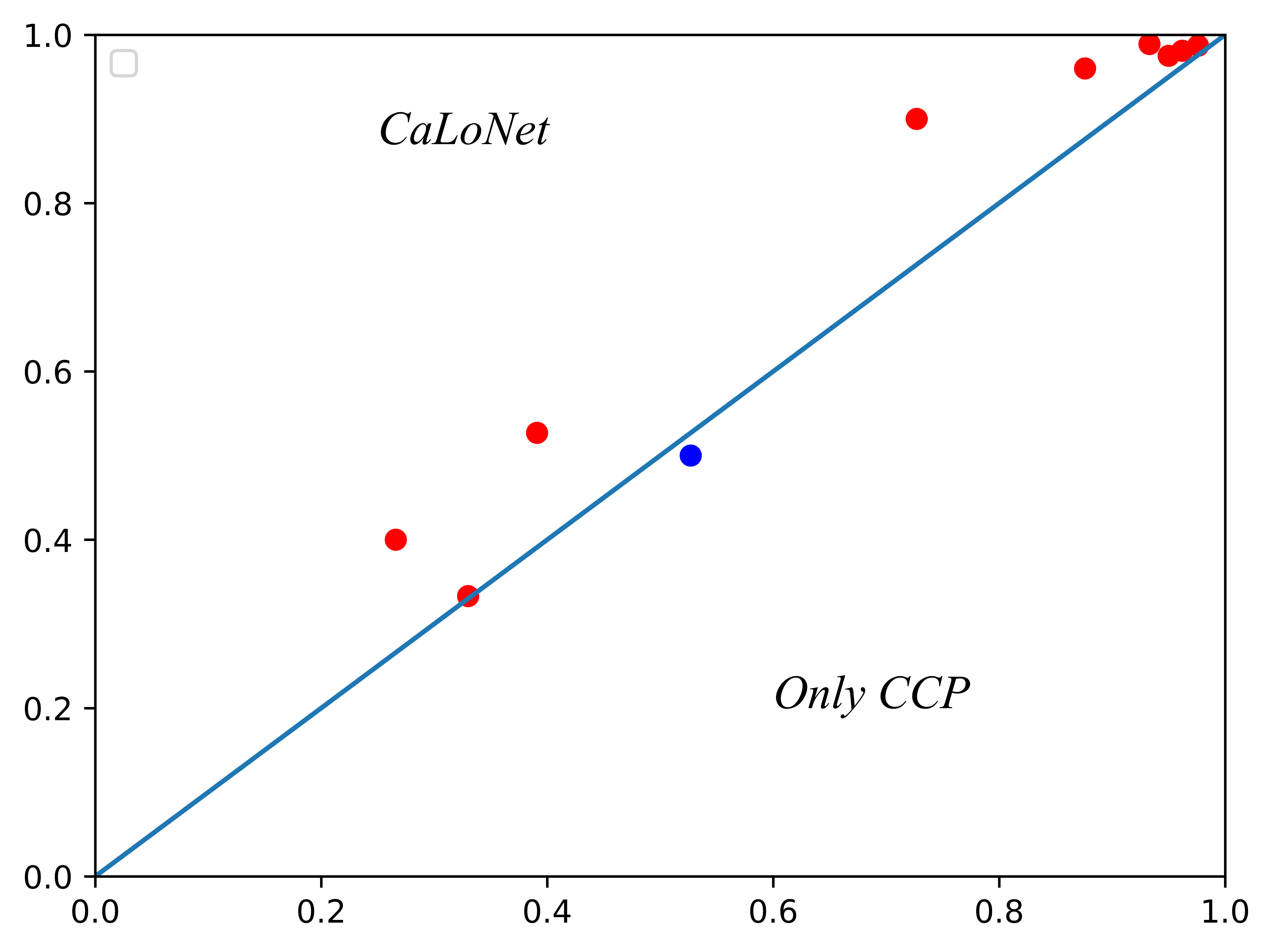}
			\label{fig:fig9b}
		}	
		\caption{Pairwise accuracy comparison between CaLoNet and Only LCP or Only CCP.}
		\label{fig:fig9}
	\end{figure}

	The causal correlations part (CCP) and local correlations part (LCP) of CaLoNet were separately removed to verify the effect of each part on the performance of the model. Table~\ref{tab3} lists the comparison results of Only CCP, Only LCP, and CaLoNet. Table~\ref{tab3} and Fig.\ref{fig:fig9a} reveal that the average accuracy of CaLoNet is higher than that of Only LCP. Table\ref{tab3} and Fig.~\ref{fig:fig9b} reveal that CaLoNet improves the average accuracy of Only CCP by 6.2\%. Overall, CaLoNet achieves 5 wins out of 10 datasets in the three sensitivity comparisons. This result demonstrates that the combined effects of causal correlations and local correlations are effective for the MTS classification task.
	
	\begin{table}[!tb]
		\centering
		\caption{Comparison of the accuracies of Only LCP, Only CCP, and CaLoNet.}
		\label{tab3}
		\footnotesize
		\begin{tabular}{llll}
			\hline
			Dataset                  & Only LCP       &  Only CCP       & CaLoNet        \\ \hline
			AWR & 0.955 & 0.876          & \textbf{0.960 }         \\
			AF        & \textbf{0.400} & 0.330          & 0.333          \\
			BM              & 0.850          & 0.950          & \textbf{0.975} \\
			CT     & 0.985          & 0.976          & \textbf{0.987} \\
			HMD     & \textbf{0.540} & 0.391          & 0.527          \\
			NA                & \textbf{0.902} & 0.727          & 0.900          \\
			PD                 & 0.979          & 0.962          & \textbf{0.983} \\
			SR2        & 0.477          & \textbf{0.527} & 0.500          \\
			SAD        & 0.982          & 0.933          & \textbf{0.989} \\
			SWJ             & \textbf{0.410} & 0.266          & 0.400          \\ \hline
			AVG                  & 0.748          & 0.693          & \textbf{0.755} \\
			Win                       & 4     & 1              & \textbf{5}      \\ \hline       
		\end{tabular}
	\end{table}
	
	\subsubsection{Epoch analysis for CaLoNet}

\begin{figure}[!tb]
		\centering
		\subfloat[BasicMotions]{
			\centering
			\includegraphics[width=0.3\textwidth]{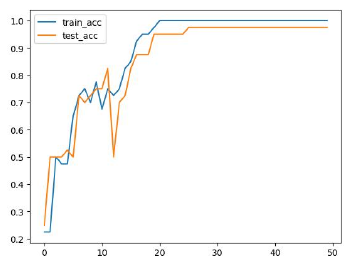}
			\label{fig:fig13a}
		}
		\subfloat[HandMovementDirection]{
			\centering
			\includegraphics[width=0.3\textwidth]{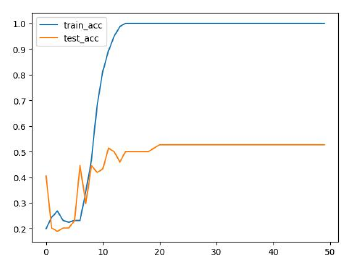}
			\label{fig:fig14q} 
		}
		\subfloat[SpokenArabicDigits]{
			\centering
			\includegraphics[width=0.3\textwidth]{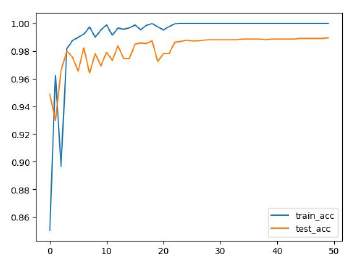}
			\label{fig:fig15a} 
		}
		\caption{Accuracy at different numbers of epochs.}
		\label{fig:fig10}
	\end{figure}

\begin{figure}[!tb]
		\centering
		\subfloat[BasicMotions]{
			\centering
			\includegraphics[width=0.3\textwidth]{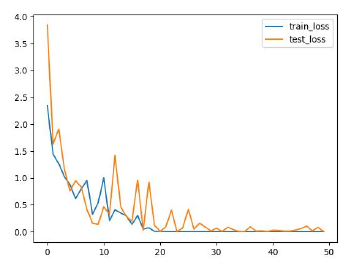}
			\label{fig:fig13b}
		}
		\subfloat[HandMovementDirection]{
			\centering
			\includegraphics[width=0.3\textwidth]{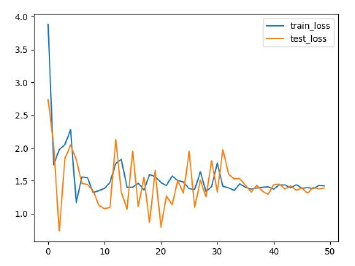}
			\label{fig:fig14b} 
		}
		\subfloat[SpokenArabicDigits]{
			\centering
			\includegraphics[width=0.3\textwidth]{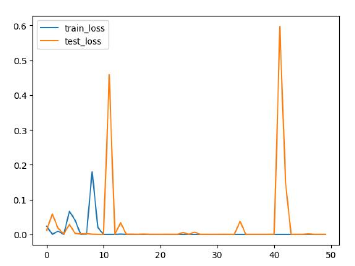}
			\label{fig:fig15b} 
		}
		\caption{Loss at different numbers of epochs.}
		\label{fig:fig11}
	\end{figure}

	To analyze the effectiveness of the number of epochs, we obtained the loss and accuracy plots for epochs ranging from 1 to 50. The accuracy and loss on the BasicMotions, HandMovementDirection, and SpokenArabicDigits datasets are shown in Figs. \ref{fig:fig10} and \ref{fig:fig11}, respectively. From these results, we observe the following:
	In the initial stage, the training and testing losses gradually decrease, while the accuracy increases as the number of epochs increases.
	When the number of epochs exceeds 30, the training and testing accuracy plateau.
	As the model is trained, the training and testing losses gradually converge, indicating the excellent fitting ability of our model.

	\subsection{ Local correlation visualization}
	
\begin{figure}[!tb]
		\centering
		\subfloat[Dimension 1 of sample 1 of ``BasicMotions"]{
			\centering
			\includegraphics[width=0.475\textwidth]{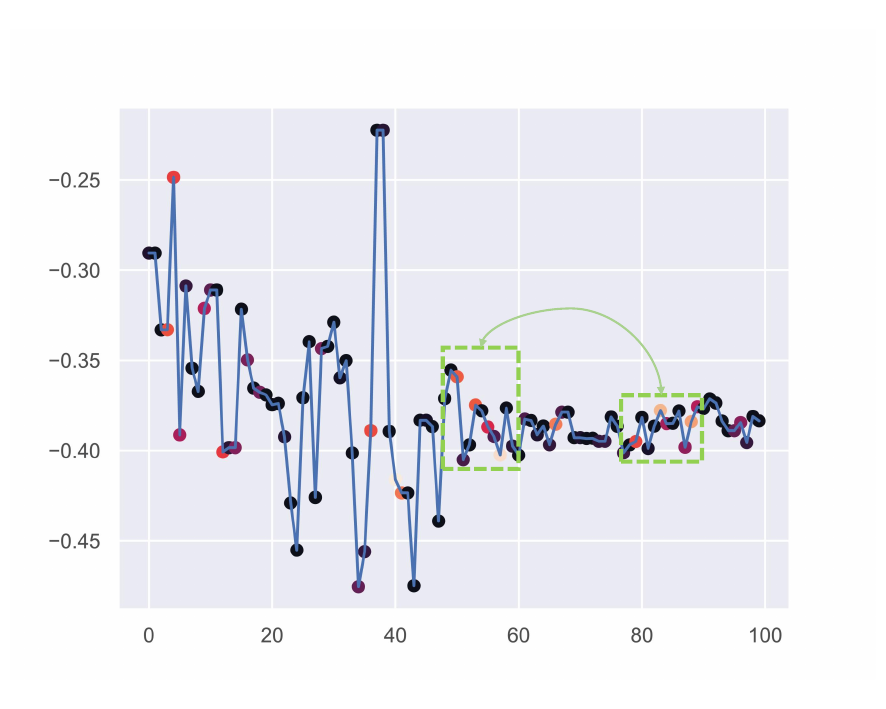}
			\label{fig:fig12a}
		}
		\subfloat[Dimension 1 of sample 2 of ``BasicMotions"]{
			\centering
			\includegraphics[width=0.475\textwidth]{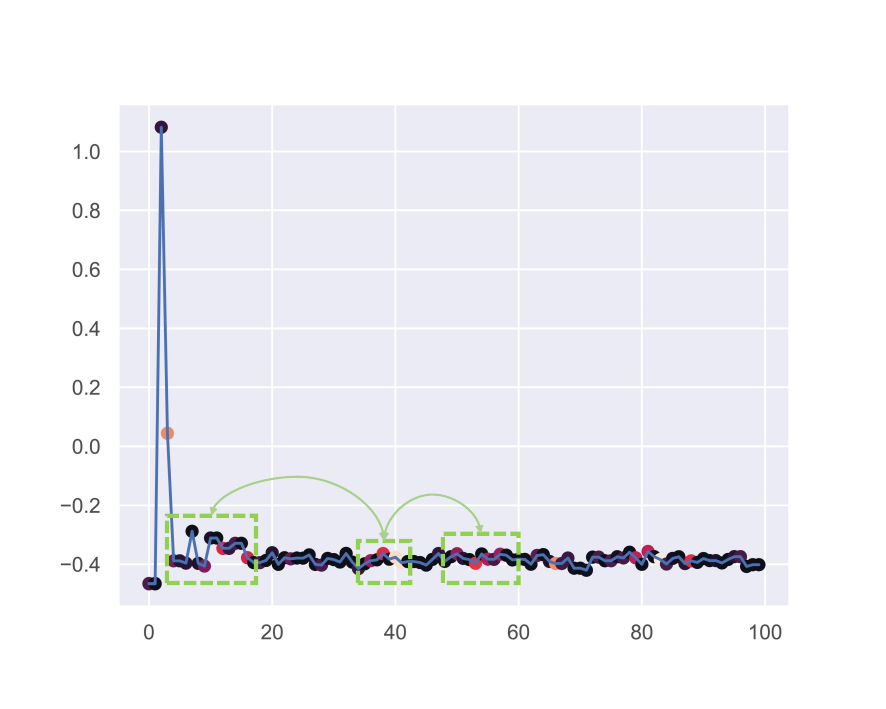}
			\label{fig:fig12b}
		}
		
		\subfloat[Dimension 1 of  sample 3 from BasicMotions]{
			\centering
			\includegraphics[width=0.475\textwidth]{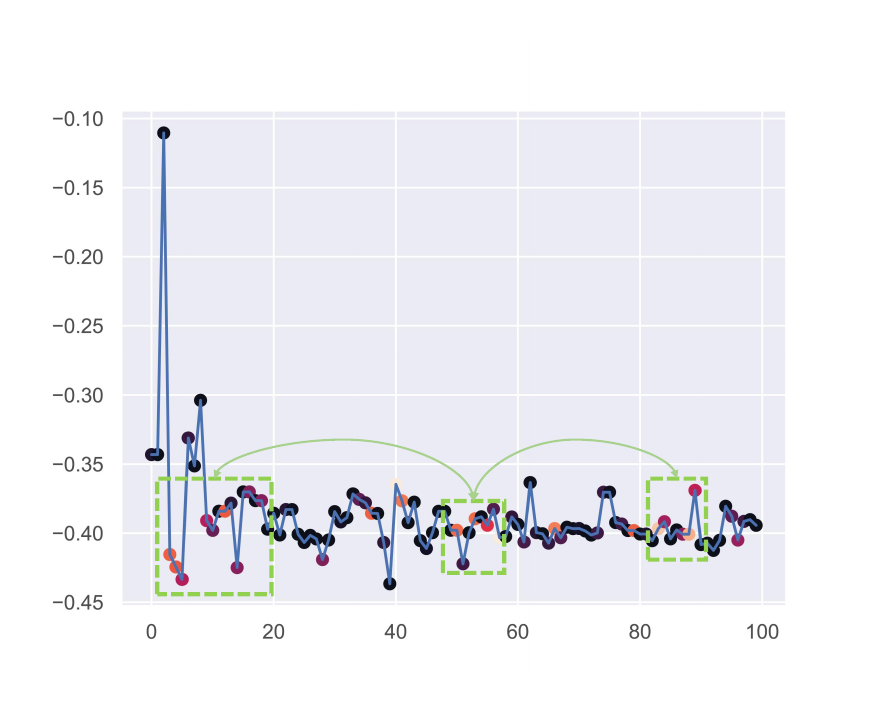}
			\label{fig:fig12c}
		}
		\subfloat[Dimension 1 of sample 4 from BasicMotions]{
			\centering
			\includegraphics[width=0.475\textwidth]{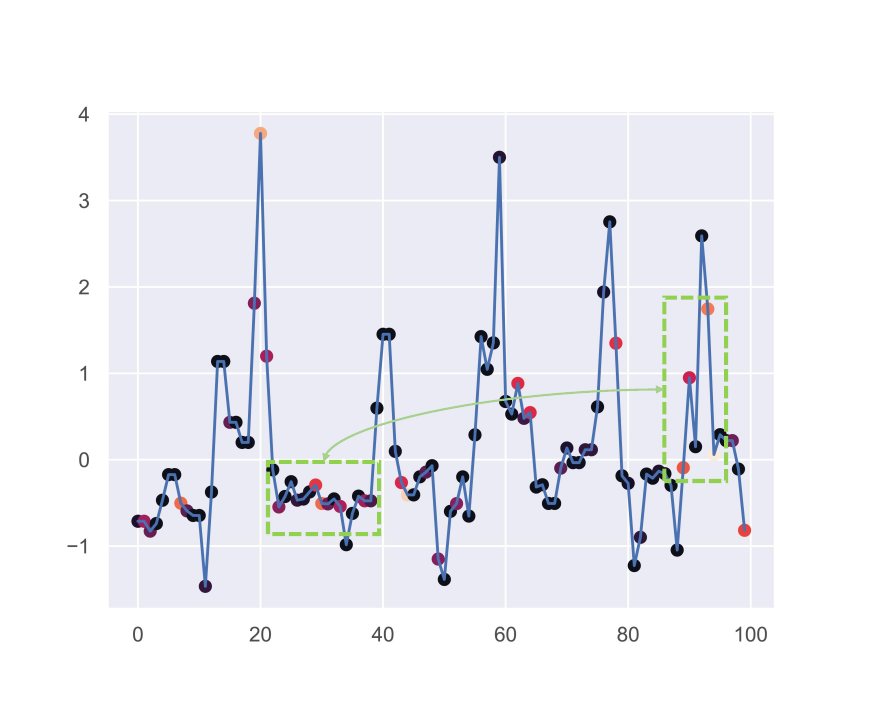}
			\label{fig:fig12d}
		}	
		\caption{Heatmap of different samples from the BasicMotions dataset.}
		\label{fig:fig12}
	\end{figure}

	We used Grad-CAM \cite{selvaraju2017grad} to visualize the local correlations of the  BasicMotions dataset. As shown in Fig.~\ref{fig:fig12}, CaLoNet captures the distinguishing local features (green dashed rectangles). This visualization illustrates the local correlation capture capabilities of CaLoNet well.

	The BasicMotions dataset was generated by four students who wore smartwatches while performing four activities: walking, resting, running, and badminton. The watches collected data from 3D accelerometers and 3D gyroscopes. Each participant recorded the activities five times, with data sampled once every tenth of a second over a ten-second period.

	The visualization in Fig.~\ref{fig:fig12} depicts real-time series data, where the horizontal axis represents the length of the time series and the vertical axis represents the corresponding values of the time series. The green connecting lines in the graph indicate that the local features within the green dashed box are closely related, meaning there is local correlation.

	Local correlation is crucial for understanding motions. It helps us identify which segments of time series in a motion contribute most significantly to its recognition, as well as the relationships between these segments. Therefore, in the BasicMotion dataset, the model extracts time series from the smartwatch data to obtain local correlations for identifying motions such as walking, resting, running, and badminton.
	
	\subsection{ Causal correlations visualization }
\begin{figure}[!tb]
		\centering
		\subfloat[HandMovementDirection]{
			\centering
			\includegraphics[width=0.32\textwidth]{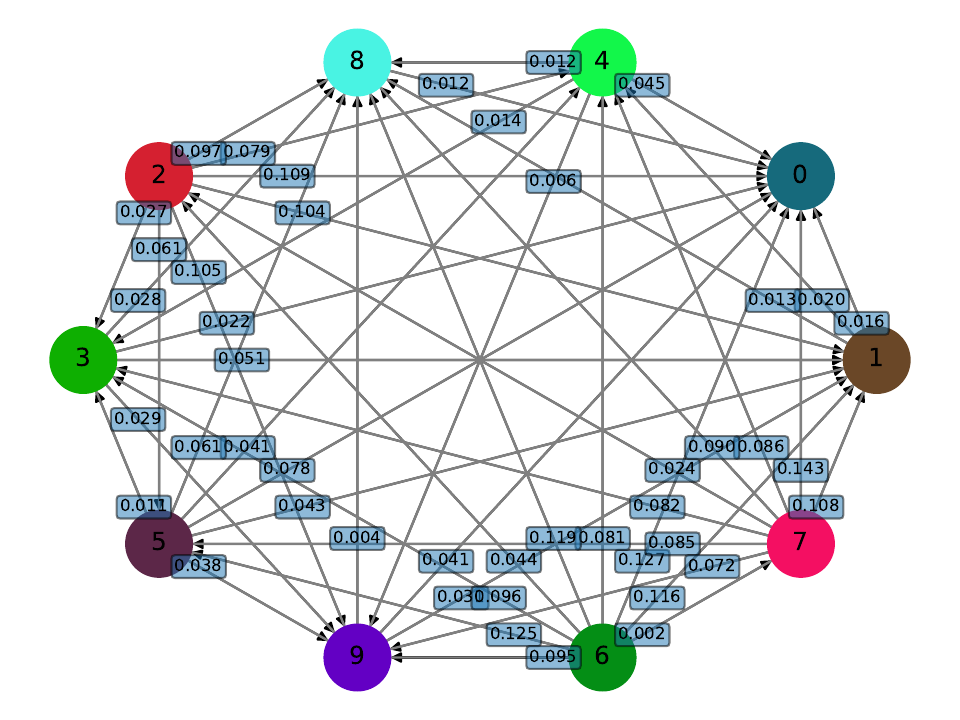}
			\label{fig:fig16a}
		}
		\subfloat[BasicMotions]{
			\centering
			\includegraphics[width=0.32\textwidth]{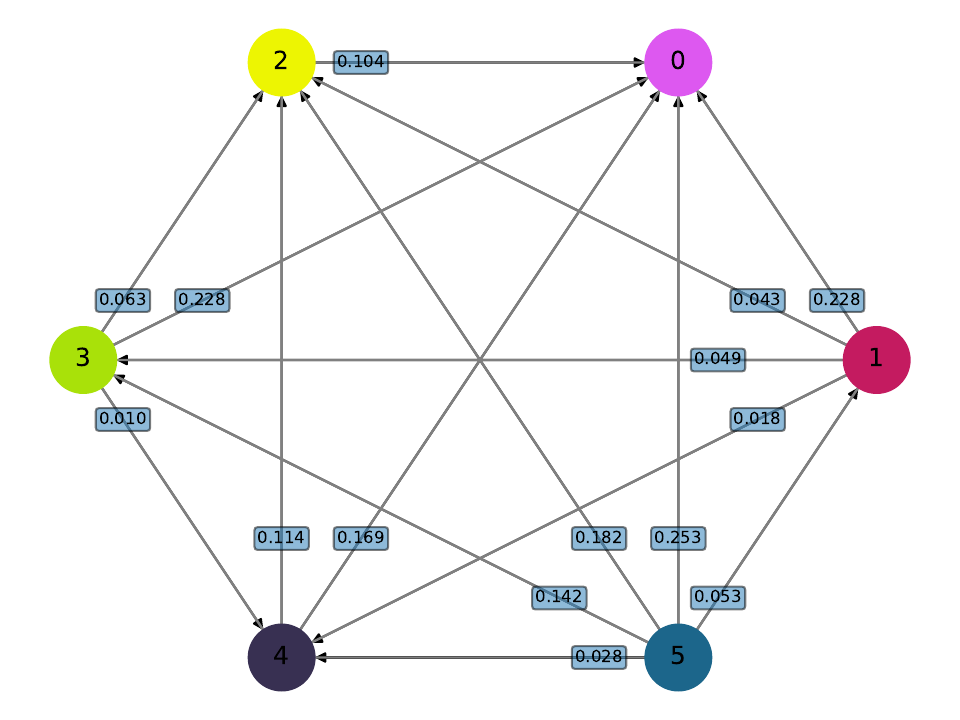}
			\label{fig:fig16b} 
		}
		\subfloat[CharacterTrajectories]{
			\centering
			\includegraphics[width=0.32\textwidth]{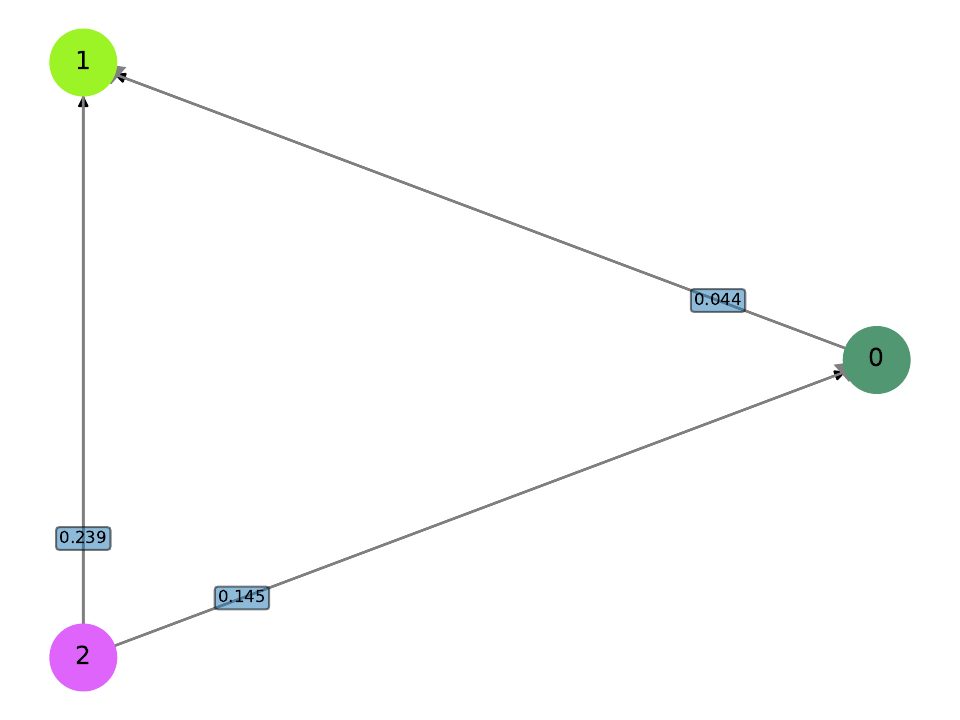}
			\label{fig:fig16c} 
		}
		\caption{Causal correlation-based graph structures from various datasets.}
		\label{fig:fig16}
	\end{figure}

	Fig.~\ref{fig:fig16} visualizes the causal correlation-based graph structures of three datasets: HandMovementDirection (10 dimensions), BasicMotions (6 dimensions), and CharacterTrajectories (3 dimensions). In these graph structures, the colored dots with arrows are the causes and those that are pointed at are the effects, and the cause and effect between each dimension form the spatial correlations.
	
	In Fig~\ref{fig:fig16}, each colored vertex represents a node, and the edges between nodes indicate causal relationships between dimensions. In addition, the parameters on the line represent the strength of the connection between the nodes. We use transfer entropy to quantify the causal relationships of MTS, aiming to discover potential causal interactions among variables. Transfer entropy can identify whether there exists a potential information transfer pathway between variable sequences, i.e., whether the past state of one variable can better predict the future state of another variable. This information transfer relationship can serve as evidence of potential causal interactions between variables.

	The CharacterTrajectories data were captured using a WACOM tablet, using three dimensions: x, y, and pen tip force. The data were captured at a frequency of 200 Hz and normalized. Only characters with a single PEN-DOWN segment were considered. Each sample is a 3D pen tip velocity trajectory. The original data had cases of different lengths, which were truncated to the shortest length (182) to conform to the repository.

\begin{figure}[!tb]
		\centering
		\subfloat[HandMovementDirection]{
			\centering
			\includegraphics[width=0.32\textwidth]{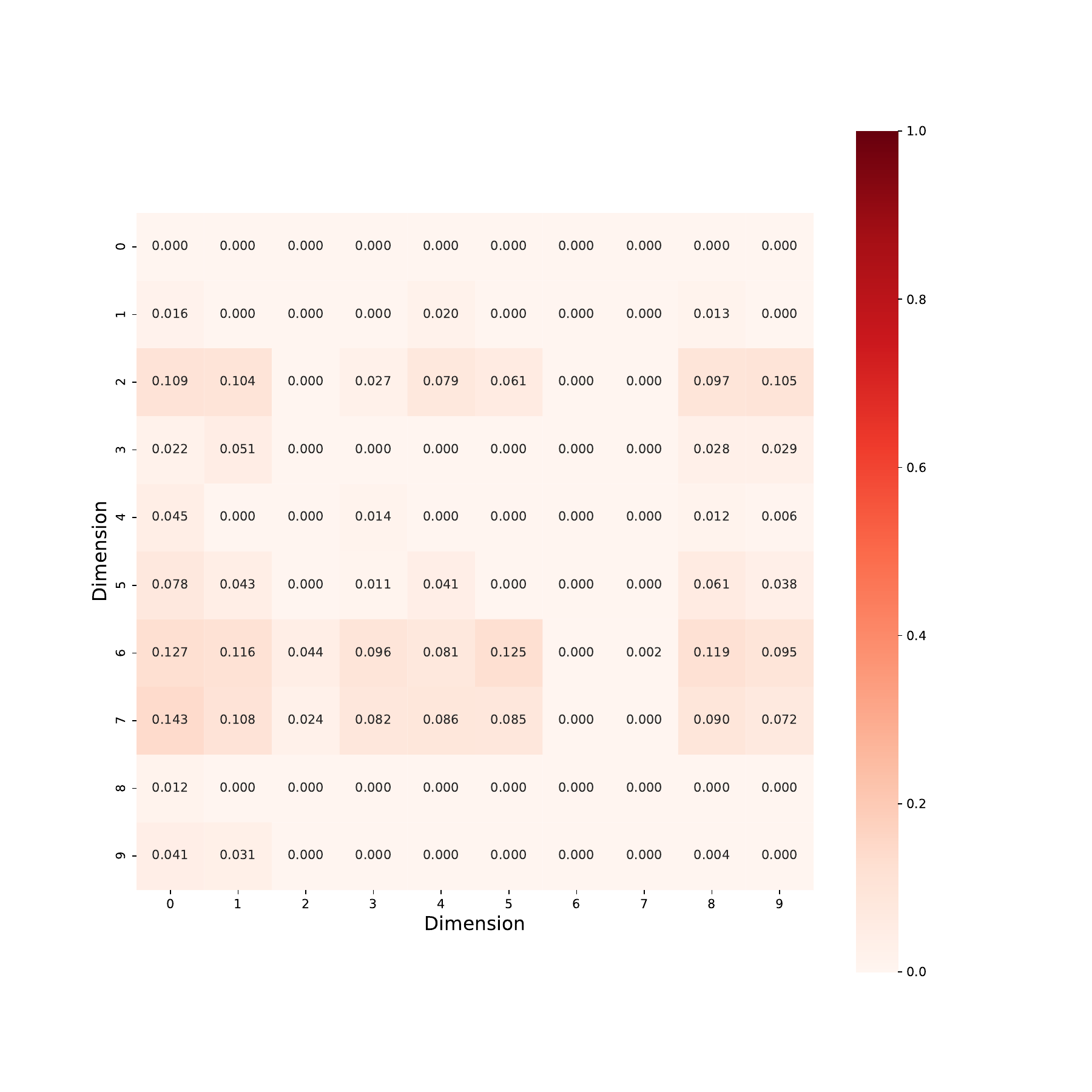}
			\label{HMD-relitu}
		}
		\subfloat[BasicMotions]{
			\centering
			\includegraphics[width=0.32\textwidth]{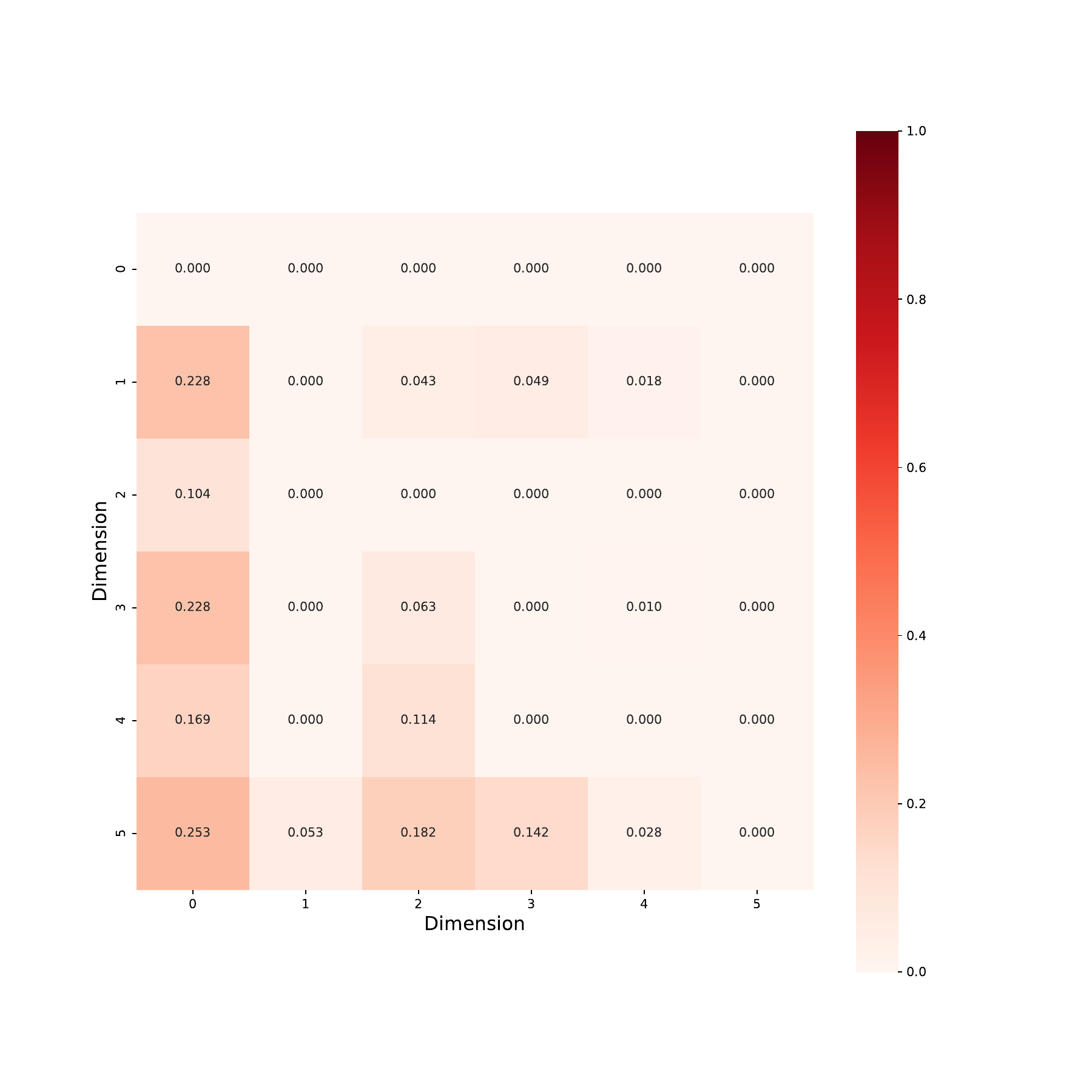}
			\label{BM-relitu} 
		}
		\subfloat[CharacterTrajectories]{
			\centering
			\includegraphics[width=0.32\textwidth]{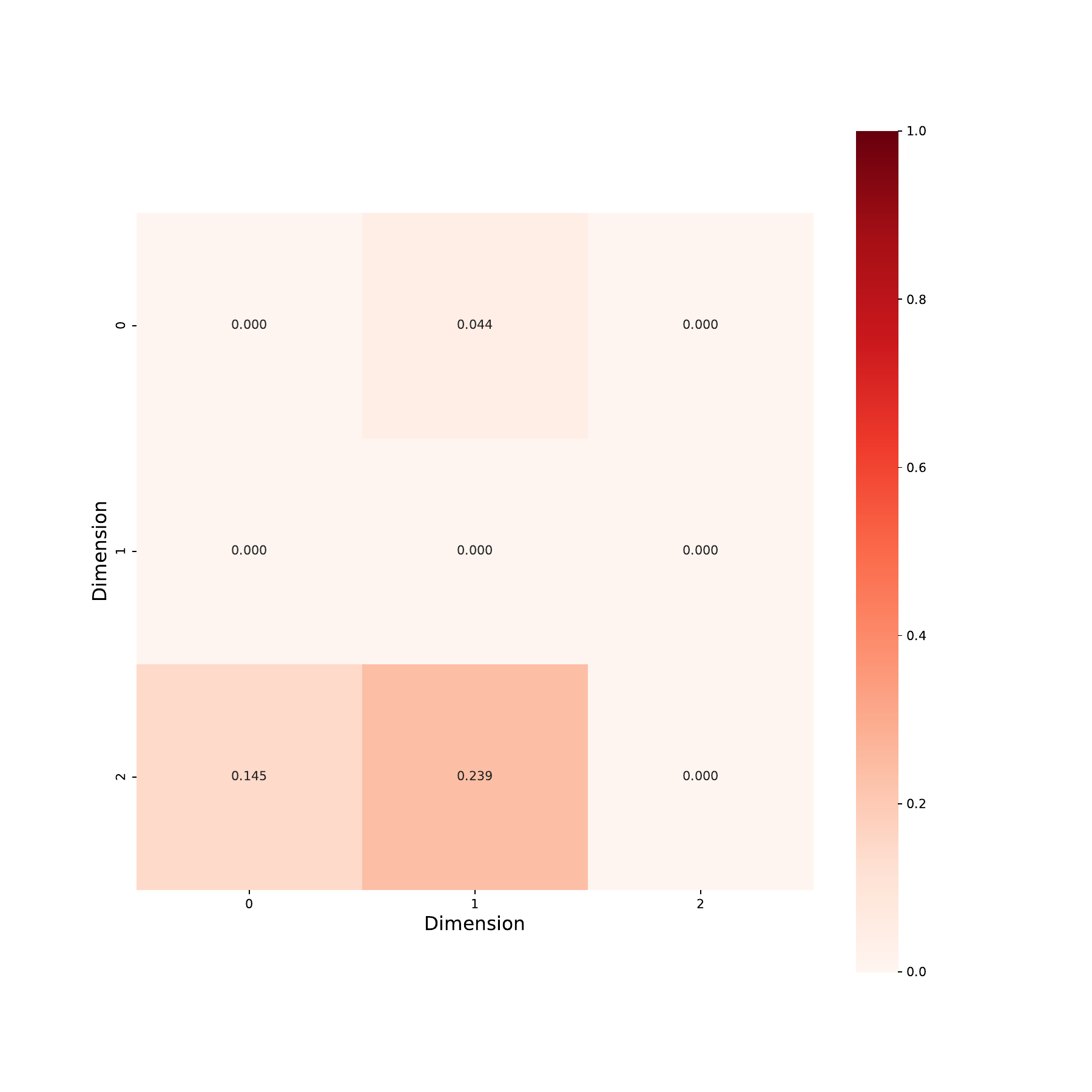}
			\label{CT-relitu} 
		}
		\caption{Heatmaps for various datasets.}
		\label{relitu}
	\end{figure}

	Fig~\ref{relitu} shows a causal relationship heatmap in which the colors represent the strength of causal influence (larger numbers in the box represent stronger causal correlations). The numerical value of transfer entropy also quantifies the strength of causal influence between different variables to a certain extent, providing a basis for modeling and analysis. For example, in the CharacterTrajectories dataset, the third dimension is the cause of the first (0.1450) and second dimensions (0.2389), whereas the first dimension is the cause of the second dimension (0.0443).
	
	\section{Conclusion}
	\label{sec:con}
	In this paper, we proposed CaLoNet, a new end-to-end deep-learning model for exploring local and causal correlations. CaLoNet first leverages pairwise spatial correlations between dimensions based on causal correlations to model and obtain the graph structure. Second, a relationship extraction network fuses features to obtain local correlations and capture long-term dependency features. Finally, the graph structure and long-term dependency features are incorporated into a GNN to complete the MTS classification task. Our experiments demonstrated an improvement in accuracy.
	
	The static graph structure used in this study is unable to capture the dynamic changes in relationships that occur as data changes over time. This is a limitation of the approach, because in many real-world applications, the interactions between variables are dynamic and evolve over time, and dynamic modeling is required to better reflect the true underlying relationships.
	To address this limitation, we plan to explore the use of a dynamic GNN in future work. Dynamic GNNs offer the ability to model the dynamic nature of graph structures by incorporating temporal information and relationship changes over time.

	\section*{Acknowledgements}
	This work was supported by the Innovation Methods Work Special Project under Grant 2020IM020100 and the Natural Science Foundation of Shandong Province under Grant ZR2020QF112. We thank Kimberly Moravec, PhD, from Liwen Bianji (Edanz) (www.liwenbianji.cn/) for editing the English text of a draft of this manuscript.  
	
	\bibliographystyle{elsarticle-num}
	\bibliography{refference}
\end{document}